\documentclass[journal]{IEEEtai}

\usepackage{xr-hyper}   

\usepackage[colorlinks,urlcolor=blue,linkcolor=blue,citecolor=blue]{hyperref}

\usepackage{color,array}
\usepackage{graphicx}
\usepackage{float}

\ifCLASSOPTIONcompsoc
    \PassOptionsToPackage{%
        caption=false,
        font=normalsize,
        labelfont=sf,
        textfont=sf
    }{subfig}
\else
    \PassOptionsToPackage{%
        caption=false,
        font=footnotesize
    }{subfig}
\fi

\def\makinhome{/home/makin}
\makeatletter
\begingroup\endlinechar=-1\relax
    \everyeof{\noexpand}%
    \edef\x{\endgroup\def\noexpand\homepath{%
        \@@input|"kpsewhich --var-value=HOME" }}\x
\makeatother

\ifx\homepath\makinhome
    \newcommand{\stydir}{../../stys}
    \newcommand{\bibsdir}{.}
    \newcommand{\tikzdir}{../../tikzpics/BSFF}

\else
    \newcommand{\stydir}{stys}
    \newcommand{\bibsdir}{bibs}
    \newcommand{\tikzdir}{tikzpics}

\fi

\providecommand{\stydir}{../stys}
\providecommand{\bibsdir}{../bibs}
\providecommand{\tikzdir}{../tikzpics}

\usepackage{%
	amsmath,amsfonts,
	amssymb,amsthm,graphicx,rotating,booktabs,
	versions,ifthen,siunitx,bbm,
}
\usepackage{xcolor}
\usepackage{pgf,tikz,pgfplots} 
\usetikzlibrary{%
 	positioning,calc,patterns,shapes,arrows,intersections,backgrounds,fit,shadings,decorations.text,shadows,fadings,bayesnet,
}
\usepgfplotslibrary{colorbrewer,colormaps,fillbetween,groupplots}

\ifdefined\rvmacroize
\else
	\makeatletter
	\input{\stydir/jgmstd.sty}
	\makeatother
\fi

\newboolean{INCLFIGS} 
\setboolean{INCLFIGS}{true}

\usepackage{subfig}
\newcommand{\captioning}[2]{\caption{{\bf #1} {#2}}}
\ifthenelse{\boolean{INCLFIGS}}{
}{
	\captionsetup[subfloat]{labelformat=empty,labelsep=none}
}

\usepackage{\stydir/IMSL}

\renewcommand{\eqn}[1]{(\ref{eqn:#1})}
\renewcommand{\eqns}[2]{(\ref{eqn:#1}) and (\ref{eqn:#2})}

\usepackage{algorithm}
\usepackage{algorithmic}

\newcommand{\appref}[1]{Appendix \ref{sec:#1}}
\renewcommand{\captioning}[2]{\caption{{#1} {#2}}}
\renewcommand{\argcolor}{black}
\renewcommand{\argcolor}{black}

\rvmacroize[*]{image}
\rvmacroize{class}
\rvmacroize{dataconditioner}
\rvmacroize[][][\argcolor]{argconditioner}
\rvmacroize[*]{real}
\rvmacroize{hiddim}

\def\ltntsym{u}
\def\obsvsym{z}
\rvmacroize[\text{in}][!]{inwt}
\def\INWT{\INWTS_\text{in}}
\rvmacroize[\text{out}][!]{outwt}

\def\OUTWT{\OUTWTS_\text{out}}
\rvmacroize[\text{out}][!]{barwt}

\renewcommand{\logisticop}[1]{\sigma\left(#1\right)}

\newcommand{\inverselink}[1]{\psi\left(#1\right)}
\def\logpartitionsym{A}
\newcommand{\logpartitionop}[1]{A\left(#1\right)}
\renewcommand{\partialarg}[1]{\partial{#1}}

\def\prior#1 {%
    \assignkeys{%
        distributions, gener, conditioner=\argconditioners{},
        paramdisplay={;\parameters}, parameters=\INWT,
        adjust, #1}%
    \distribution{\latent|\conditioner\paramdisplay}
}
\def\priorfactor#1 {%
    \assignkeys{%
        distributions, gener, index=\hiddim, latent=\argltnt{\index}, conditioner=\argconditioners{},
        paramdisplay={;\parameters}, parameters=\inwt{\index},
        adjust, #1}%
    \distribution{\latent|\conditioner\paramdisplay}
}
\def\marginal#1 {%
    \assignkeys{%
        distributions, gener, conditioner=\argconditioners{}, paramdisplay={;\parameters}, parameters={\INWT,\OUTWT}, adjust, #1%
    }%
    \distribution{\patent|\conditioner\paramdisplay}%
}
\def\emission#1 {%
    \assignkeys{%
        distributions, gener, conditioner=\argconditioners{}, paramdisplay={;\parameters}, parameters=\OUTWT, adjust, #1%
    }%
    \distribution{\patent|\latent\paramdisplay}%
}
\def\posterior#1 {%
    \assignkeys{%
        distributions, gener, conditioner=\argconditioners{}, paramdisplay={;\parameters}, parameters={\OUTWT,\INWT}, adjust, #1%
    }%
    \distribution{\latent|\patent,\conditioner\paramdisplay}%
}
\def\joint#1 {%
    \assignkeys{%
        distributions, gener, conditioner=\argconditioners{}, paramdisplay={;\parameters}, parameters={\OUTWT,\INWT}, adjust, #1%
    }%
    \distribution{\patent,\latent|\conditioner\paramdisplay}%
}

\def\generpriorfactor#1 {\priorfactor #1 }

\newcommand{\FigForwardForwardPGM}{
    \begin{figure}[!t]
        \scriptsize
        \tikzstyle{factor IPGM}=[factor, fill=none, draw=black]%
        \subfloat[][]{
            \label{subfig:logisticFF}
            \begin{tikzpicture}

                \node[obs] (v) {$\Dataconditioners$};

                \node[obs, above=1in of v] (x) {$\Dataobsvs$};
                \node[above left=0.15in and -1.2in of x] (xlabel){$
                    \genermarginal{paramdisplay={}}
                        \propto
                    h(\argobsvs)\expop{\argobsvs\tr\OUTWT\logisticop{\INWT\Dataconditioners}}%
                $};

                \edge[->] {v} {x};

                \plate{allvars}{(x)(v)}{$N$};
            \end{tikzpicture}
        }
        \subfloat[][]{
            \label{subfig:BSFF}
            \begin{tikzpicture}
                \node[obs] (v) {$\Dataconditioners$};

                \node[latent, above=0.36in of v] (u) {$\Dataltnts$};
                \node[overlay, left=0.6in of u, text width=0.5in, align=center] (ulabel){$%
                    \generprior{paramdisplay={}} 
                        =
                    \prod_k^K \Bern{\logisticop{\inwts{k}\cdot\argconditioners}}%
                $};

                \node[obs, above=0.36in of u] (z) {$\Dataobsvs$};
                \node[above=0.1in of z] (zlabel){$%
                    \generemission{paramdisplay={}}
                        \propto
                    h(\argobsvs)\expop{\argobsvs\tr\OUTWT\argltnts}
                $};

                \edge[->] {v} {u};
                \edge[->] {u} {z};

                \plate{allvars}{(u)(v)(z)}{$N$};
            \end{tikzpicture}
        }
        \captioning{%
            Discriminative models for \subcapref{logisticFF} deterministic and \subcapref{BSFF} binary-stochastic forward-forward.
        }{In Hinton's FF, $\protect\Dataconditioners$ corresponds to labeled images and $\protect\Dataobsv{}$ to indicates whether the label is good or bad; in CwC-FF, $\protect\Dataconditioners$ is the image and $\protect\Dataobsvs$ the one-hot label.}
    \end{figure}
}

\begin{document}

\title{Energy-Efficient Supervised Learning with a\\ Binary Stochastic Forward-Forward Algorithm}



\author{
    Risi Jaiswal, Supriyo Datta, and Joseph G.\ Makin\\
    Elmore School of Electrical and Computer Engineering, Purdue University
}

\maketitle

\begin{abstract}
Reducing energy consumption has become a pressing need for modern machine learning, which has achieved many of its most impressive results by scaling to larger and more energy-consumptive neural networks.
Unfortunately, the main algorithm for training such networks, backpropagation, poses significant challenges for custom hardware accelerators, due to both its serial dependencies and the memory footprint needed to store forward activations for the backward pass.
Alternatives to backprop, although less effective, do exist; here the main computational bottleneck becomes matrix multiplication.
In this study, we derive forward-forward algorithms for binary, stochastic units.
Binarization of the activations transforms matrix multiplications into indexing operations, which can be executed efficiently in hardware.
Stochasticity, combined with tied weights across units with different biases, bypasses the information bottleneck imposed by binary units.
Furthermore, although slow and expensive in traditional hardware, binary sampling that is very fast can be implemented cheaply with p-bits (probabilistic bits), novel devices made up of unstable magnets.
We evaluate our proposed algorithms on the MNIST, Fashion-MNIST, and CIFAR-10 datasets, showing that its performance is close to real-valued forward-forward, but with an estimated energy savings of about one order of magnitude.
\end{abstract}

\begin{IEEEImpStatement}
Over the last decade, neural networks have become ubiquitous in scientific, industrial, and commercial applications, but suffer from a major defect: they consume large amounts of energy, both during training and at run time.
The resources needed to train AI systems have been doubling approximately every 3.4 months over this period, which is likely unsustainable, at least without major environmental costs.
Here we propose an algorithm for deep learning that maintains high accuracy on moderately difficult benchmark image-processing tasks, but with a projected energy consumption approximately a tenth that of standard training algorithms.
Our algorithm replaces multiplications with binary operations and minimizes data movement, allowing for operation on ultra-low-power hardware.
We demonstrate performance on GPUs, and calculate from first principles the expected energy costs on dedicated hardware.
We anticipate that these results will motivate the fabrication of a new type of AI chip, bringing on-device learning to everyday devices.
\end{IEEEImpStatement}


\section{Introduction}\label{sec:intro}
\IEEEPARstart{I}{n} 
recent years, deep neural networks (DNNs) have revolutionized natural language processing, computer vision, automatic speech recognition, and various scientific domains \cite{Devlin2018,krizhevsky2012imagenet,Hannun2014,Jumper2021}.
Tasks that seemed exceedingly complex even 15 years ago are now feasible, and have commercially deployed solutions.
However, a major component of this progress has been the use of complex models that require substantial computational resources, typically in the form of high-powered GPU clusters.
In fact, the computational resources needed to train AI systems have been doubling approximately every 3.4 months since 2012 \cite{economist2020cost}.
It is not clear that the energy demands required to sustain this pace of development can be met, especially given concerns about anthropogenic climate change \cite{motherjones2024energy,MIT2024energy}.

While there has been a considerable emphasis on accelerating inference methods \cite{Howard2017,li2019fully,Ankit2019,gholami2022survey,Hubara2018} to run on low-energy devices, enhancements in training methodologies have not kept pace.
This is because nearly all DNNs are trained with the backpropagation algorithm \cite{Rumelhart1986}, which presents two major obstacles to implementation on efficient custom hardware.
First, backprop requires the unit activations computed in the forward pass to be stored (for use in computing the gradient in the backward pass).
This imposes a significant memory cost.
Furthermore, although the forward pass can be computed efficiently in analog hardware \cite{Aguirre2024,Tsai2018,Hu2016,Shafiee2016}, 
storage in memory requires a digital representation, and analog-to-digital conversion dominates the silicon area and power consumption in these methods \cite{Tsai2018,Shafiee2016}.
Second, the forward passes cannot be run asynchronously or on pipelined data, but must be interrupted by weight updates that traverse the full depth of the network.

Motivated in part by these considerations, there has been recent interest in developing alternatives to backprop \cite{Frenkel2021,Hinton2022,Kohan2022,Dellaferrera2022,jiang2023one}.
For example, Hinton has recently proposed an algorithm for supervised learning, called ``forward-forward,'' that uses only forward passes through the DNN.
Although less effective than backprop, forward-forward avoids both of the aforementioned obstacles to implementing backprop in hardware.

Nevertheless, there is another obstacle to implementing neural networks in energy-efficient, analog hardware:\ noise accumulation.
Analog/mixed-signal neurons are more susceptible to degradation in classification accuracy as bitwidth increases compared to their digital counterparts \cite{Chatterjee2019}.
This requires careful tuning of design to make it work.
A potential solution is to quantize the activations or the weights into fewer bits \cite{gholami2022survey,Courbariaux2014}.
At the extreme, activations could even be binarized \cite{Qin2020,Peters2018}.
This is appealing because matrix multiplications against binary vectors can be carried out as indexing and adding operations, which can be executed very efficiently in hardware \cite{zhao2017accelerating,nurvitadhi2016accelerating,zhou2017deep}.

Moreover, some of the precision lost to binarization can be recouped by \emph{sampling} from neurons.
For example, a pair of binary, stochastic neurons reporting (let us suppose) the same feature will together provide (on average) more information than their deterministic counterparts.
Setting aside the issue of matrix multiplications, this scheme might seem far less efficient than simply quantizing less severely.
But in fact a recent hardware innovation called the \emph{p-bit} has made it possible to draw Bernoulli-distributed voltages at very high speeds, low power consumption, and small footprint \cite{Kaiser2019}.
We discuss the p-bit below (\sctn{pbit}).

Taken together, these considerations point toward a new, more energy-efficient approach to supervised learning, based on (1) foward-pass-only algorithms rather than backprop; (2) binary, stochastic neurons; and (3) implementation in hardware with p-bits.
To demonstrate the feasibility of this approach, in this work we extend to binary, stochastic activations both Hinton's original forward-forward algorithm \cite{Hinton2022} and a more powerful recent extension \cite{papachristodoulou2024convolutional}.
We demonstrate that our binary stochastic forward-forward algorithms achieve accuracy comparable to their real-valued counterparts.
Although we implement our algorithms in software, we estimate that the hardware implementation would save approximately one to two orders of magnitude in energy costs.

\section{Background}
\subsection{Alternatives to backpropagation}
Our work combines two distinct research directions, one focused on non-backpropagation-based algorithms and the other on propagating gradients through binary stochastic neurons.
The motivation behind non-backpropagation methods for training deep neural networks primarily arises from the biological implausibility of backpropagation.
Many key challenges for biological implementations---weight transport, the need for symmetric weights, non-local computation, propagating global error signals from the output to every layer---likewise pose difficulties for efficient hardware implementations.
However, certain other biologically implausible features, like weight sharing in convolutional layers, can enhance hardware performance by saving memory and improving energy efficiency in digital circuits.

Feedback Alignment (FA) \cite{Lillicrap2016} addresses the weight transport problem by using random feedback weights, eliminating the need for symmetric weights.
Similarly, Direct Feedback Alignment (DFA) \cite{Nokland2016} directly uses error signals from output layers with random feedback weights instead of sequentially propagating.
Although FA and DFA circumvent the problems of weight symmetry and weight transport, errors still need to propagate through sequential connection in FA and through direct connection in DFA.
This poses a problem for asynchronous continuous learning on input data, which limits some of the potential benefits for hardware efficiency.

The forward-forward (FF) algorithm proposed by Hinton \cite{Hinton2022} replaces the traditional forward and backward passes with two forward passes using positive and negative examples, thus making asynchronous updates possible. 
A recent extension of FF, channelwise-competitive FF (CwC-FF) \cite{papachristodoulou2024convolutional}, replaces the contrastive examples with competitive learning across channels, and achieves higher accuracies.
These are the algorithms we extend in the present study (see \sctn{theory} below).
Cascaded Forward (CaFo) \cite{zhao2023cascaded} is similar in spirit, with layerwise losses, only forward passes, and no contrastive examples, but for want of space we do not explore it here.
The recently introduced PEPITA \cite{Dellaferrera2022} can also be interpreted as two forward passes, the first with the input data, and the second with these data adjusted based on direct error feedback.
But accuracy compares unfavorably with FF.

There exists a distinct literature on biologically plausible learning in energy-based models, most notably Equilibrium Propagation \cite{Scellier2017}.
This approach, in which weight updates can be seen as as an approximation of backpropagation, has shown potential for efficient implementation in energy-agnostic physical systems, as further explored in \cite{Scellier2022}.

\subsection{Binary stochastic neurons}
The other body of work that overlaps with this investigation focuses on propagating gradients through binary stochastic neurons (BSNs)---first clearly addressed by Bengio and colleagues \cite{Bengio2013}.
Following this paper, the ``straight-through estimator'' (STE) has become the \emph{de facto} method for training neural networks with BSN; more precisely, the use of the identity function in place of the gradient through samples from a BSN.

In this study, we are concerned with the basic question of whether p-bits can be used to port alternatives to backprop to efficient hardware.
Accordingly, we restrict our attention to the forward-forward algorithm, including its most recent variant \cite{papachristodoulou2024convolutional}, and leave application to other alternatives to backprop to future work.
We propose two estimators:\ \emph{binary stochastic forward-forward}, which is derived along lines similar to the \cite{Bengio2013,Shekhovtsov2020}; and \emph{stochastic-gradient forward-forward}, in which the gradient calculation is approximated using the samples computed in the forward passes.

\subsection{Probabilistic Bits}\label{sec:pbit}
The probabilistic bit, or \emph{p-bit}, is a concept used in probabilistic computing, an emerging paradigm that exploits physical randomness to perform computations \cite{Sutton2020,kaiser2021probabilistic,Aadit2021}.
Unlike classical bits, which can be in one of two definite states (0 or 1), p-bits take on values of 0 or 1 with probabilities determined by the device's inputs.
This probabilistic behavior can be harnessed for various computational tasks such as machine learning, optimization, and probabilistic logic.

These p-bits can be realized in various ways, such as using magnetic tunnel junctions (MTJs) in series with transistors \cite{Camsari2017}, as experimentally demonstrated by Borders and colleagues \cite{Borders2019}.
Such a \emph{p-bit} is a three-terminal device that generates a stream of ones and zeros, with the probability given by the hyperbolic tangent (tanh) of the input. The response of p-bits can be described mathematically as
\begin{align}\label{eqn:pbit}
    V_\text{out} = \text{sgn}[\tanh(V_\text{in}) - r],
    && 
    r \sim \unif{-1}{1}.
\end{align}
Here, $V_\text{in}$ is an \emph{analog} input voltage, and $V_\text{out}$ is a \emph{binary} output voltage.
This characteristic is popularly referred to as a binary stochastic neuron (BSN) in the machine learning literature.
These devices can be used to implement BSNs in neural network circuits besides its application in other domains.

\section{Theory}\label{sec:theory}
The aim of the forward-forward algorithm is to train neural networks to solve supervised-learning problems---prototypically, mapping inputs $\Images$ to class labels $\Class$---without the use of backpropagation \cite{Hinton2022}.
The central idea is to construct a loss that can be applied independently and greedily at each layer (i.e., not backpropagating to previous layers) while still driving the entire network toward a solution to the original classification problem.
In effect, forward-forward fits a sequence of neural networks with one hidden layer, rather than a single neural network with many hidden layers.

In the classical treatment of neural networks with one hidden layer, the output depends on two weight matrices, input-to-hidden ($\INWT$) and hidden-to-output ($\OUTWT$); and biologically dubious ``weight transport'' is still required, since the updates to $\INWT$ depend on $\OUTWT$.
To avoid this, forward-forward treats $\OUTWT$ as a set of fixed parameters rather than learnable weights, so the dependence in the updates is likewise fixed for all time, and no weight transport is required.
In Hinton's original version of the algorithm, the output matrix is set to be a row vector,
\begin{equation}\label{eqn:HintonReadoutWeights}
    \OUTWT
        \defeqleft
    \frac{1}{\Hiddim}\ones\tr
\end{equation}
(with $\Hiddim$ the dimension of the hidden layer), which transforms the activations in the hidden layer into the natural parameters for a \emph{binary classification problem}.
In particular, ``good'' examples are constructed by pairing images with their correct labels, ``bad'' examples by pairing images with the wrong labels.
A network capable of distinguishing good from bad classes has implicitly solved the original, multiway classification problem because it knows which labels belong with which images.
Slightly more formally, let the random variable $\Dataconditioners$ be an (image, label) pairing, and $\Dataobsv{}$ the indicator of ``good'' or ``bad'':
\begin{align*}
    \Dataconditioners
        :=
    (\Images, \Class),
    &&
    \Dataobsv{}
        = 
    \begin{cases}
        1 & \text{if $\Class$ is the correct label for $\Images$}\\
        0 & \text{otherwise.}
    \end{cases}
\end{align*}
In the standard fashion, we attempt to model the true conditional probability $\dataemission{patent=\argobsv{},latent=\argconditioners} $ with a parameterized model, 
$\genermarginal{patent=\argobsv{}} $, by minimizing their (conditional) relative entropy.
(We use circumflexes to indicate models throughout.)
Since learning requires a forward pass with good data and a forward pass with bad data (rather than a forward and a backward pass), Hinton dubbed this the ``forward-forward'' algorithm.

Alternatively \cite{papachristodoulou2024convolutional}, we can require each layer to solve the original multiway classification problem.
We still need to keep the output matrix fixed, but this time it must transform hidden-layer activities into the natural parameters for a categorical, rather than Bernoulli, distribution.
Therefore we let the output matrix be
\begin{equation}\label{eqn:GreekReadoutWeights}
    \OUTWT
        \defeqleft
    \frac{1}{\Hiddim/\Ncat}
    \begin{bmatrix}
        \ones   & \zeros    & \cdots    & \zeros    & \zeros\\
        \zeros  & \ones     & \cdots    & \zeros    & \zeros\\
        &       & \ddots    & \\        & \\
        \zeros  & \zeros    & \cdots    & \zeros    & \ones\\
    \end{bmatrix}\tr.
\end{equation}
Each of the $\Ncat$ rows of $\OUTWT$ corresponds to a category.
Thus it ``assigns'' subsets of the activities to different categories and then simply averages activities within these subsets.
With this modification, it is no longer necessary to construct new input data, because it is no longer necessary to turn the multiway classification problem into a binary classification.
In this case, the input is just the image, $\Dataconditioners = \Images$, and the output $\Dataobsvs$ is a one-hot class label.
We follow Papachristodoulou and colleagues in calling their algorithm channelwise-competitive FF (CwC-FF)---although this is something of a misnomer since there is no longer any need for a second forward pass with ``bad'' examples.

These forward-forward algorithms can be extended to multiple hidden layers by treating one hidden layer's activities as inputs to the next.
However, the activities must first be normalized, so as to hide this layer's classification from succeeding layers:\ only the features, but not their magnitudes, are passed on.
Since this is the only requirement, we provide (next) the mathematical descriptions only for two-layer networks.

\FigForwardForwardPGM

\subsection{Logistic Forward-Forward}
Before deriving binary stochastic forward-forward algorithms, we begin with the real-valued, deterministic version, in particular two-layer neural networks with logistic (sigmoidal) units $\reals$.
In order to cover both Bernoulli and categorical random variables, we write the output distribution as a generic exponential family (suppressing all biases for concision),
\begin{equation}\label{eqn:logisticForward}
    \begin{split}
        \real{\hiddim}
            &=
        \logisticop{\inwts{\hiddim}\cdot\dataconditioners},
        \qquad \hiddim \in [1, \Hiddim]\\
        \genermarginal{}
            &=
        h(\argobsvs)\expop{\argobsvs\tr\ntrlparams{}(\reals) - \logpartitionop{\ntrlparams{}(\reals)}},
    \end{split}
\end{equation}
where $\inwts{\hiddim}$ is row $\hiddim$ of $\INWT$, the natural parameters $\ntrlparams{} \defeqleft \OUTWT\reals$ are linear in the hidden-layer activities, and $\logisticop{\cdot}$ is the logistic (sigmoid) function.
Our objective is to minimize the (conditional) relative entropy of the data and model:
\begin{align}
    \label{eqn:marginalRelativeEntropy}
    \mathcal{L}
        &\defeqleft
    \relativeentropy{patent/\Dataobsvs,conditioner/\Dataconditioners}{\datamarginal}{\genermarginal}\\
        &=
    \def\integrand#1 {
        \assignkeys{distributions, gener, adjust, #1}%
        \logpartitionop{\ntrlparams{}(\reals)} - \patent\tr\ntrlparams{}(\reals)
    }
    \sampleaverage{patent/\Dataobsvs,conditioner/\Dataconditioners}{\integrand} + C,    
\end{align}
with $C$ a constant independent of the parameters.
The gradient of this loss with respect to $\inwts{l}$, a single row of $\INWT$, takes on the (familiar) form
\def\observations#1 {%
    \assignkeys{distributions, gener, adjust, #1}%
    \patent
}
\begin{equation}\label{eqn:LFFgradient}
    \begin{split}
        \colttlderiv{\mathcal{L}}{\inwts{l}}
            &=
        \def\integrand#1 {%
            \assignkeys{distributions, gener, adjust, #1}%
            \left(\colgradient{\logpartitionsym}{\ntrlparams{}}(\ntrlparams{}(\reals)) - \patent\right)\tr
            \colgradient{\ntrlparams{}}{\reals}
            \colgradient{\reals}{\inwts{l}}
        }
        \sampleaverage{patent/\Dataobsvs,conditioner/\Dataconditioners}{\integrand}\\
            &=
         \def\integrand#1 {
            \assignkeys{distributions, gener, adjust, #1}%
            \left(
                \inverselink{\OUTWT\reals} - \patent
            \right)\tr
            \outwts{l}
            \real{l}(1 - \real{l})
            \conditioner
        }
        \sampleaverage{patent/\Dataobsvs,conditioner/\Dataconditioners}{\integrand},
    \end{split}
\end{equation}
with $\outwts{l}$ the $\lth$ column of $\OUTWT$.
On the second line we have used the fact that, for exponential families, the gradient of the log-partition function is the expectation of the sufficient statistics, in this case $\Generobsvs$; or, equivalently, the inverse link (logistic for Bernoulli outputs, the softmax for categorical, etc.), which we refer to here as $\inverselink{\cdot}$, applied to the natural parameters.

Ordinarily, \eqn{LFFgradient} is interpreted as propagating the output gradient,
$ \inverselink{\OUTWT\reals} - \Dataobsvs$,
backwards through the output weights $\outwts{l}$, which raises questions for biological plausibility (how could the synaptic weights in the feedback path be locked to the feedforward weights?) and hardware implementations (see \sctn{intro}).
In the forward-forward algorithm, however, the elements of $\OUTWT{}$ are fixed and equal (or zero; see \eqns{HintonReadoutWeights}{GreekReadoutWeights}), so $\outwts{l}$ can be absorbed into the gradient step size for all $l$.

\subsection{Binary Stochastic Forward-Forward}
To port this algorithm efficiently to hardware, we need to replace the matrix multiplications with indexing and adding.
This can be achieved if we make the hidden units binary- rather than real-valued.
Simply thresholding the outputs, $\reals$, of the first layer in \eqn{logisticForward} discards useful information; it is preferable to draw samples, which across multiple training inputs provides real-valued information.
Conveniently, drawing Bernoulli samples is exceedingly cheap under the proposed hardware platform (\sctn{pbit}).
Therefore, let the real values $\reals$ specify the firing probabilities, rather than the activities, of random hidden units $\Generltnts$:
\begin{equation}\label{eqn:independentBernoullis}
    \begin{split}
        \generprior{}
            &=
        \prod_{\hiddim=1}^{\Hiddim} \generpriorfactor{} \\
            &=
        \prod_{\hiddim=1}^{\Hiddim} \logisticop{\inwts{\hiddim}\cdot\argconditioners}^{\argltnt{\hiddim}}
        \left(1 - \logisticop{\inwts{\hiddim}\cdot\argconditioners}\right)^{1-\argltnt{\hiddim}}\\
        \generemission{}
            &=
        h(\argobsvs)\expop{\argobsvs\tr\ntrlparams{}(\argltnts) - \logpartitionop{\ntrlparams{}(\argltnts)}},
    \end{split}
\end{equation}
with the natural parameters now linear in the binary samples, $\ntrlparams{} = \OUTWT\argltnts$, rather than $\reals$.

To fit this model, we ought to minimize the same relative entropy as in the deterministic case, \eqn{marginalRelativeEntropy}---but in the stochastic case, this equation hides the fact that the latent Bernoulli random variables $\Generltnts$ have been marginalized out.
The gradient likewise takes a superficially simple form, but in practice introduces an intractable normalizer.
Although this can be finessed with an importance sampler (see \appref{losses}), we achieve superior results by instead descending the gradient of a ``variational'' upper bound.
In particular, we consider the loss in \eqn{marginalRelativeEntropy} to be a ``marginal'' relative entropy, and then follow the classic variational approach \cite{Neal1998} in constructing the following ``joint'' relative entropy (suppressing parameters for brevity):
\pgfkeys{distributions, adjust/.style={/distributions/.cd, paramdisplay={}}}%
\begin{equation}\label{eqn:jointRelativeEntropy}
    \begin{split}
        \mathcal{J} 
            &\defeqleft
        \def\integranda#1 {\datamarginal{#1} \generprior{#1} }
        \relativeentropy{latent/\Generltnts,conditioner/\Dataconditioners,patent=\Dataobsvs}{\integranda}{\generjoint}\\
            &=
        \def\integranda#1 {
            \generprior{#1,paramdisplay={}}
            \log\generemission{#1,paramdisplay={}}
        }
        \def\integrandb#1 {
            \dmarginalize{latent/\generltnts}{\integranda #1,}
        }
        \sampleaverage{patent/\Dataobsvs,conditioner/\Dataconditioners}{\integrandb} + C.
    \end{split}
\end{equation}
This is an upper bound on the relative entropy in \eqn{marginalRelativeEntropy} because it can be expressed as the sum of that relative entropy and a non-negative quantity (another relative entropy):
\begin{equation*}
    \mathcal{J}
        =
    \relativeentropy{latent/\Generltnts,conditioner/\Dataconditioners,patent=\Dataobsvs}{\datamarginal}{\genermarginal}
        +
    \relativeentropy{latent/\Generltnts,conditioner/\Dataconditioners,patent=\Dataobsvs}{\generprior}{\generposterior}.
\end{equation*}
(See \appref{losses} for a fuller derivation.)
We also observe that the upper bound can be expected to be fairly tight:\ The input $\Dataconditioners$ should provide many more bits of information about $\Generltnts$ than $\Dataobsvs$, which provides only $\log{\Ncat}$, with $\Ncat$ the number of categories.
In the case of Hinton's forward-forward algorithm, this is just one bit.
So the second relative entropy should be small.

The gradient of this loss (in particular, the last line of \eqn{jointRelativeEntropy}) is tractable, and can be written (see \appref{losses}) as
\begin{equation}\label{eqn:BSFFvariationalGradient}
    \begin{split}
        \colttlderiv{\mathcal{J}}{\inwts{l}}
            &\approx
        \def\integrandb#1 {%
            \assignkeys{distributions, gener, adjust, #1}%
            \inverselink{\OUTWT\latent}
        }
        \def\integrandc#1 {
            \assignkeys{distributions, gener, adjust, #1}%
            \left(
                \condexpectation{latent/\Generltnts}{\conditioner}{\integrandb #1,}{foo/\conditioner}
            - \patent
            \right)\tr\outwts{l}
            \real{l}(1 - \real{l})\conditioner
        }
        \sampleaverage{patent/\Dataobsvs,conditioner/\Dataconditioners}{\integrandc}.
    \end{split}
\end{equation}
Note the general resemblance between \eqn{BSFFvariationalGradient} and \eqn{LFFgradient}.
Thus, the only change in the learning rule required by the move from the deterministic to the stochastic network is to replace the model class probabilities, $\inverselink{\OUTWT\reals}$, with the \emph{expected} class probabilities, 
\def\integrand#1 {%
    \assignkeys{distributions, gener, adjust, #1}
    \inverselink{\OUTWT\latent}
}%
$\condexpectation{latent/\Generltnts}{\Dataconditioners}{\integrand}{patent/\Dataconditioners}$---although we recall that \eqn{BSFFvariationalGradient} is the gradient of an upper bound to the loss, rather than the loss itself.

\section{Methods}
\subsection{Approximating the BSFF gradient}
In practice, we approximate the conditional expectation in \eqn{BSFFvariationalGradient} with a single sample vector, $\generltnts$, from the hidden layer for each input/output pair ($\dataconditioners, \dataobsvs$); hence
\begin{equation}\label{eqn:BSFFvariationalGradientB}
    \colttlderiv{\mathcal{J}}{\inwts{l}}
        \approx
    \def\integrandc#1 {
        \assignkeys{distributions, gener, adjust, #1}%
        \left(
            \inverselink{\OUTWT\latent} - \patent
        \right)\tr\outwts{l}
        \real{l}(1 - \real{l})\conditioner
    }
    \sampleaverage{patent/\Dataobsvs,latent/\Generltnts,conditioner/\Dataconditioners}{\integrandc}.
\end{equation}
We refer to the descent of \eqn{BSFFvariationalGradientB} as \textbf{binary stochastic forward-forward (BSFF)}, and consider it to be the main contribution of this study.

Still, from the perspective of hardware acceleration, \eqn{BSFFvariationalGradientB} has one flaw:\ the factor $\real{l}(1 - \real{l})$ is a real number.
Observe that this term corresponds to a smooth hill of activity centered on 0 and monotonically decreasing to zero away from 0 (blue curve, \subfig{logisticDerivative}).

We can approximate that hill with the following function:
\begin{equation}\label{eqn:stochasticHill}
    \left[\real{l} \le \frac{1}{2}\right]\generltnt{l}
        +
    \left[\real{l} > \frac{1}{2}\right](1 - \generltnt{l}),
\end{equation}
with $[P]$ equal to 1 if $P$ is true, otherwise 0 (that is, an Iverson bracket).
Unlike $\real{l}(1 - \real{l})$, this quantity is integer-valued.
Nevertheless, its expected value as a function of the input to the unit,
\begin{equation*}
    \left[\real{l} \le \frac{1}{2}\right]\real{l}
        +
    \left[\real{l} > \frac{1}{2}\right](1 - \real{l}),
\end{equation*}
shown in green in \subfig{logisticDerivative}, is close to  $\real{l}(1 - \real{l})$.
Intuitively, \eqn{stochasticHill} reports ``surprising'' events:
It is 1 when either unit $l$ ``spiked'' despite its probability of spiking being below 1/2; or when this unit did \emph{not} spike, despite having a probability of spiking \emph{above} 0.5.
Otherwise, the function is 0.
Such ``unexpected'' pairings clearly become more common as the probability of spiking approaches 1/2 from either side, or equivalently, as the input (including the bias) approaches zero.
Thus the expected value of \eqn{stochasticHill} is, like $\real{l}(1 - \real{l})$, a hill of activity peaking at zero input. 

\begin{figure}[ht]
    \centering
    \scriptsize
    \providecommand{\figwidth}{0.53\columnwidth}
    \providecommand{\figheight}{\figwidth}
    \subfloat[][]{
        \label{subfig:logisticDerivative}
        \input{\tikzdir/approxderiv}
    }
    \subfloat[][]{
        \label{subfig:stochasticSoftPlus}
        \input{\tikzdir/stochasticSoftplus}
    }
    \captioning{Approximation functions.}{%
    \subcapref{logisticDerivative} Exact (blue) and expected approximate (green) derivatives of the logistic function.
    \subcapref{stochasticSoftPlus} The response functions for an exact rectified linear unit (blue) and ``tiled logistic units,'' with one (orange), two (green), three (red), and seven (purple) tied units.
    Each tiled-logistic response function corresponds to a single realization.
    }
    
\end{figure}

Substituting \eqn{stochasticHill} for $\real{l}(1 - \real{l})$ in \eqn{BSFFvariationalGradientB} yields
\begin{equation}\label{eqn:BGBSFFgradient}
    \begin{split}
        \colttlderiv{\mathcal{J}}{\inwts{l}}
            \approx&
        \bigg\langle\left(
            \inverselink{\OUTWT\Generltnts} - \Dataobsvs
        \right)\tr\outwts{l}\\
        &
        \:\:\:\left(
            \left[\real{l} \le \frac{1}{2}\right]\real{l}
                +
            \left[\real{l} > \frac{1}{2}\right](1 - \real{l}),
        \right)
        \Dataconditioners\bigg\rangle_{\Dataobsvs,\Generltnts,\Dataconditioners}
    \end{split}
\end{equation}
We call the algorithm that descends this gradient \textbf{binary gradient, binary stochastic forward-forward (BGBSFF)}.
The energy savings from using BGBSFF over BSFF are small (because the Jacobian of the activation function is diagonal; see \appref{energyConsumption}), but we investigate the performance of both in our results below.

\subsection{Approximating ReLUs}\label{sec:approximatingReLUs}
The activation functions in the most accurate neural-network classifiers are unbounded, rather than logistic, as in our model (\eqn{independentBernoullis}).
Nevertheless, although p-bits naturally provide Bernoulli samples (recall \eqn{pbit}) and therefore a stochastic binarization of logistic units, they can be extended to stochastic quantization of other activation functions.
In particular, recall that the softplus is the integral of the logistic function. Therefore a (finite) sum over logistic functions with linearly spaced biases approximates the softplus, or its piecewise-linear counterpart, the rectified linear unit.
We therefore introduce \emph{tiled logistic units},
\begin{equation}
	\generltnt{d}
		\defeqleft
	\sum_{m=1}^{M} \left[
        \logisticop{x - m + 0.5} \ge r_m
    \right],
\end{equation}
with $x$ the input to this unit, $r_m \sim \unif{0}{1}$, and the bracket equal to 1 if its argument is true and zero otherwise (as in \eqn{stochasticHill}).
Clearly such units increase the cost of computation; we factor these into the total costs of the algorithm in \appref{energyConsumption}.

\begin{table*}[!ht]
    \begin{minipage}{\textwidth}
    \centering
    \captioning{Comparison of memory access and multiplications across algorithms.}{$N = $ batch size; $C =$ number of channels; $C_\text{in}=$ number of input channels; $H, W = $image dimensions; $L = $ number of layers.
    Only the dominant terms are shown in the table; detailed counts of multiplications and memory accesses are provided in \appref{energyConsumption}.}
    \label{tbl:energyCost}
    \begin{tabular}{@{}ccc@{}}
        \toprule
        \textbf{Method} & \textbf{Number of 32-bit memory accesses} & \textbf{Number of 32-bit multiplications} \\ 
        \midrule
        backprop & $NC^2HWL$ & $ 3NC^2K^2HWL$ \\
        CwC-FF & $NC^2HWL$  & $2NC^2K^2HWL$ \\
        BSFF & $ NC_\text{in} C HW + \frac{1}{32}NC^2HW(L-1)$
        & $2NC_\text{in} C K^2HW+4NC^2K^2(L-1)$\\ 
        \bottomrule
    \end{tabular}
    \end{minipage}
\end{table*}

\subsection{More complex layers}
Higher classification accuracies can be achieved with more complex architectures, comprising most obviously convolutions but also pooling and batch normalization \cite{Ioffe2015}.
In our binary stochastic version of the CwC-FF algorithm introduced by Papachristodoulou and colleagues \cite{papachristodoulou2024convolutional}, we incorporate these elements exactly as in their architecture.
However, batch normalization turns binary values into real numbers, and thereby indexing operations (back) into multiplication during the gradient computation.
This reduces some of the energy savings achievable with indexing.
Therefore, below we also consider computing the layerwise loss \emph{before} the batch normalization (after the max-pooling).

\subsection{Evaluation and training details}\label{sec:details}
To evaluate the performance of binary stochastic forward-forward, we tested it on three canonical image-classification datasets:\ MNIST~\cite{deng2012mnist}, Fashion-MNIST (FMNIST)~\cite{xiao2017fashionmnist}, and CIFAR-10~\cite{krizhevsky2009learning}.
We achieved the best results with the Convolutional Channel-wise Competitive Forward-Forward (CwC-FF) architecture \cite{papachristodoulou2024convolutional} (as opposed to the original forward-forward algorithm) and therefore present only those results here (see \sctn{results}).
CwC-FF uses the weight matrix in \eqn{GreekReadoutWeights} and $\inverselink{\cdot} = \text{softmax}(\cdot)$ for the inverse link function.
That is, CwC-FF divides the output channels at each layer into 10 subsets, corresponding to the number of classes.
Each subset's output is averaged to produce the natural parameter for the cross-entropy loss.

Following Papachristodoulou and colleagues \cite{papachristodoulou2024convolutional}, we compute the entire network's predictions by training a single-layer classifier on the outputs of the final layer.
All accuracies reported in \sctn{results} are from this classifier.

We evaluate performance for CwC-FF as well as both our binary-stochastic algorithms, BSFF \eqn{BSFFvariationalGradient} and BGBSFF \eqn{BGBSFFgradient}.
For the latter two, we consider logistic units as well as tiled logistic units, with either 2, 3, or 7 tiled units.
For each of these architectures as well as for the original real-valued, deterministic CwC-FF, we trained 5 models from scratch, and report the resulting distributions of classification accuracies on the test set.

We follow Papachristodoulou and colleagues in staggering the termination of training across layers.
That is, all layers are trained simultaneously, but training in layer 1 terminates first (after which the weights are frozen), followed by layer 2, etc.
When reproducing their results, i.e.\ with ReLU activation functions, we use the same termination epochs reported in that study \cite{papachristodoulou2024convolutional}.
For BSN-based activation functions, we have approximately doubled the number of training epochs for each layer.
We did not fine-tune this parameter for optimal accuracy similar to other hyperparameters.

Indeed, since the object of this study is to reduce the energy costs of training, it is crucial that the results not depend strongly on highly optimized hyperparameters, since this optimization incurs its own energy costs.
To ensure this, \emph{we did not perform hyperparameter tuning}, and instead used default values (learning rate = 0.001 and batch size = 128) for both BSFF and BGBSFF, except for the BSN:1 case, where we tuned the learning rate for the MNIST and FMNIST datasets---without which performance was quite variable.
For real-valued, deterministic CwC-FF, our main point of comparison, we used the hyperparameters reported in the original study \cite{papachristodoulou2024convolutional}, which were found by optimization, so as to provide the most conservative estimate of the relative performance of our algorithms. Detailed listings of the hyperparameters and a comparison with backpropagation can be found in Appendix~\ref{sec:supplementary}.

\section{Results}\label{sec:results}

\begin{figure*}[t]
    \centering
    \includegraphics[width=0.32\textwidth]{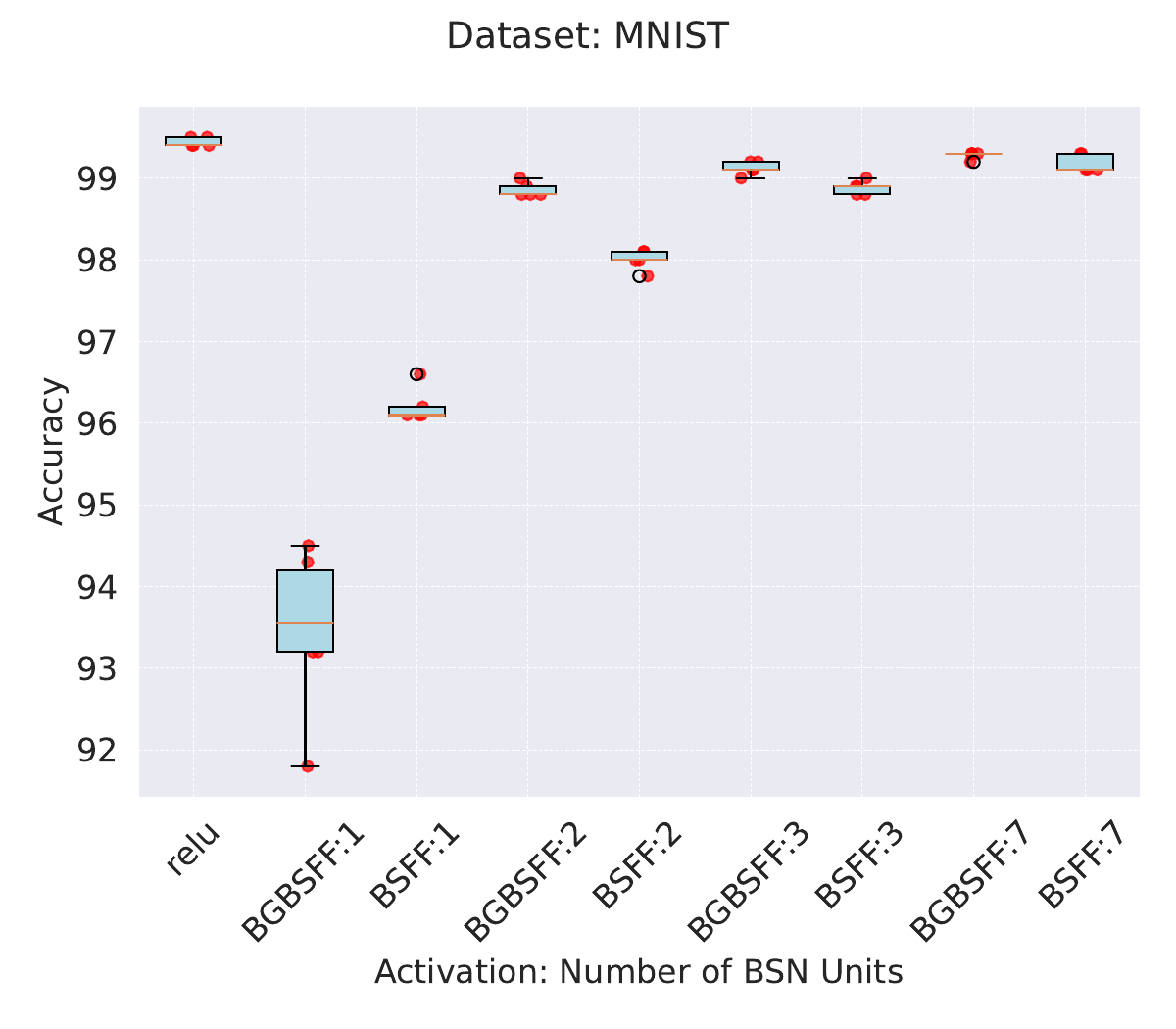} \hfill
    \includegraphics[width=0.32\textwidth]{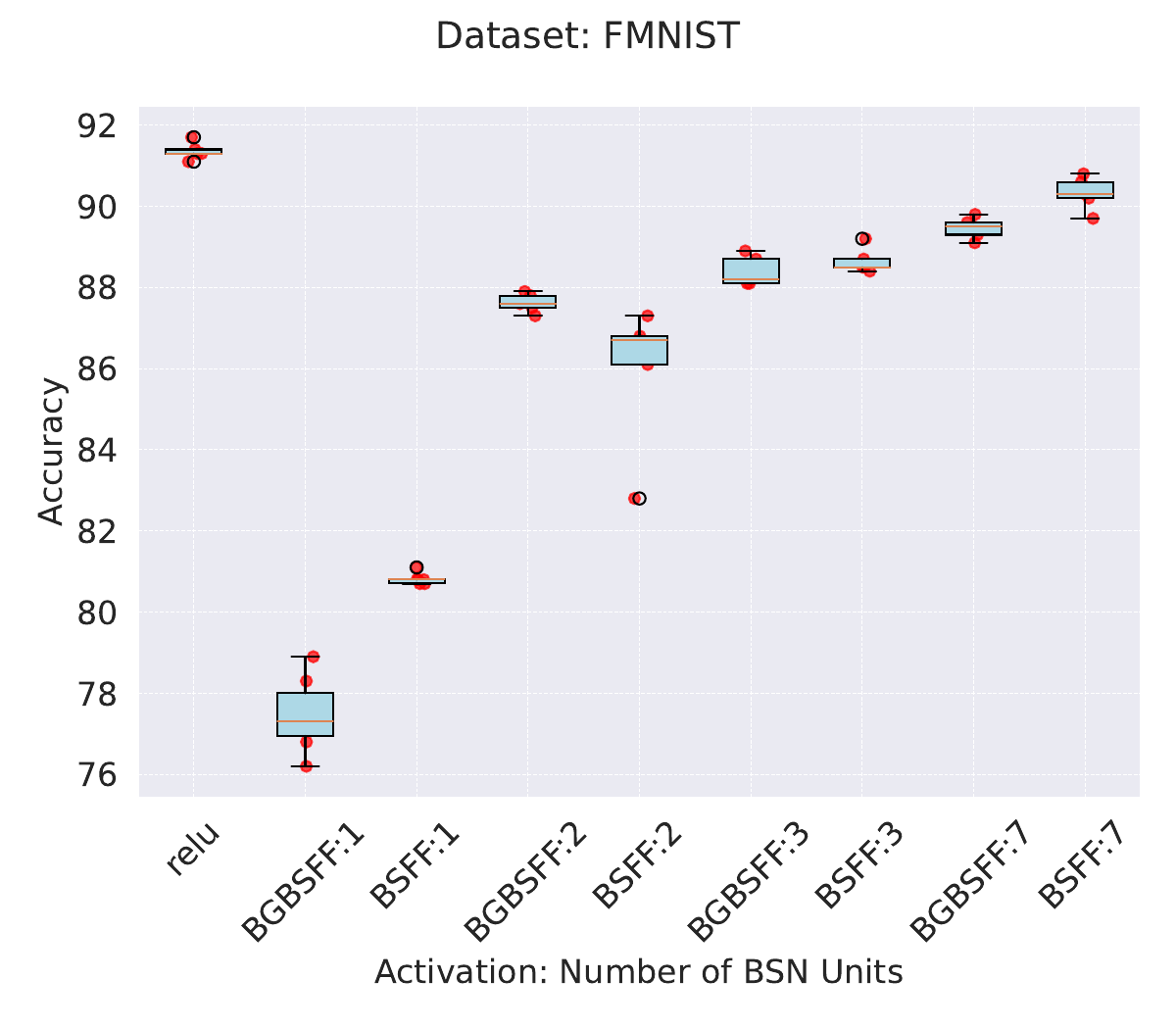} \hfill
    \includegraphics[width=0.32\textwidth]{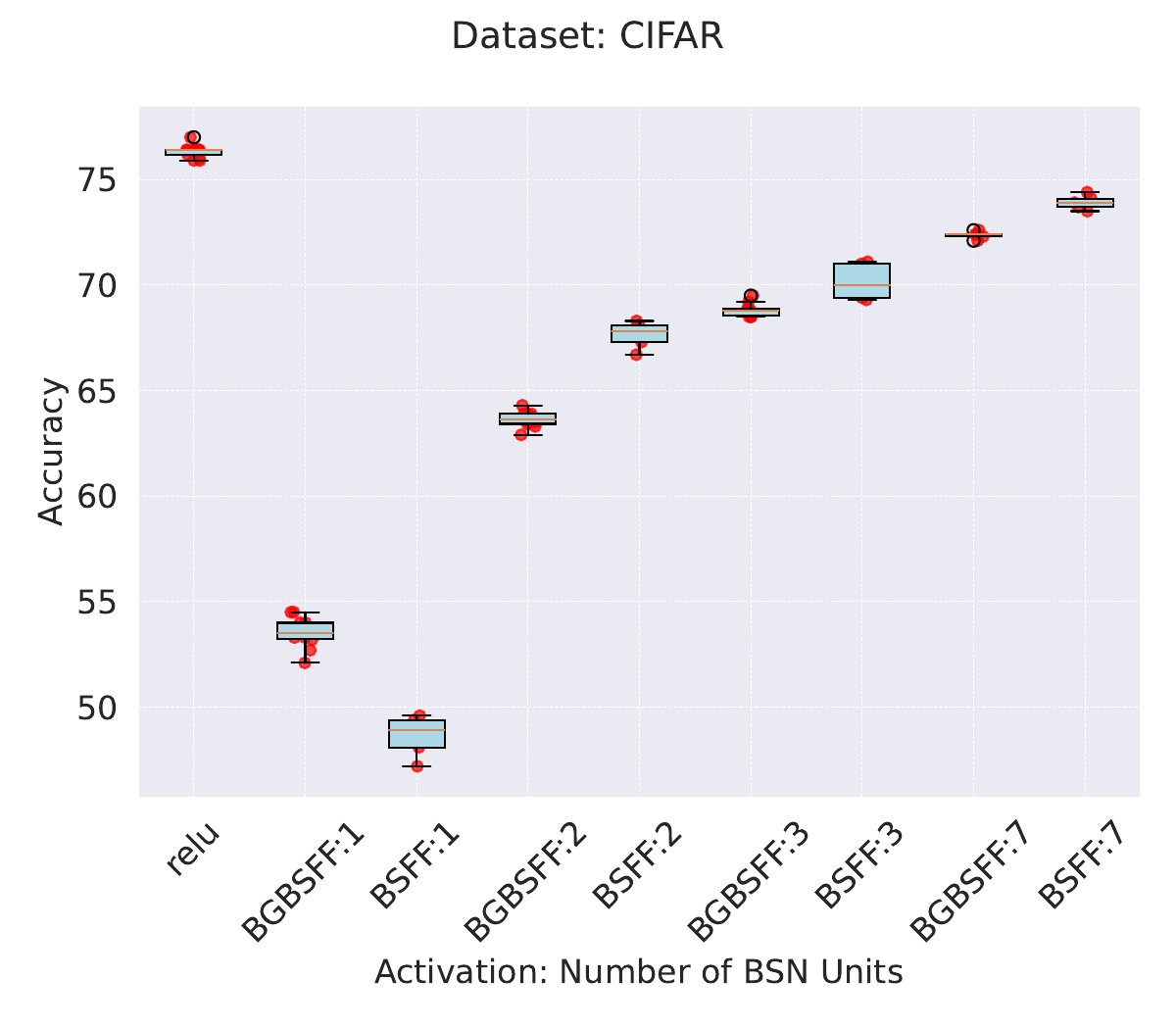}
 \captioning{Comparison of accuracies for the MNIST, FMNIST, and CIFAR-10 datasets.}{The leftmost boxplot represents results with ReLU activation, i.e.\ real-valued, deternimistic CwC-FF; while the remaining boxplots correspond to binary activations using BSFF and BGBSFF gradients, with the number of tied binary stochastic units set to 1, 2, 3, and 7, respectively.
 The box represents the inter-quartile range (IQR: Q1 to Q3), red line indicates the median (Q2), and the whiskers extend to data within 1.5$\times$IQR, while outliers are shown as individual points outside the whisker.}
    \label{fig:accuracies}
\end{figure*}

\begin{table*}[t]
    \centering
    \caption{BGBSFF with and without batch normalization}
    \label{tbl:lossAtMaxpool}
    \setlength{\tabcolsep}{3.5pt} 
    \begin{tabular}{@{}ccccccccc@{}}
        \toprule
       \textbf{Dataset} & \multicolumn{2}{c}{\textbf{BGBSFF:1}} & \multicolumn{2}{c}{\textbf{BGBSFF:2}}  & \multicolumn{2}{c}{\textbf{BGBSFF:3}} & \multicolumn{2}{c}{\textbf{BGBSFF:7}} \\
        \midrule
        &  w/BatchNorm & w/o &  w/BatchNorm & w/o &  w/BatchNorm & w/o &  w/BatchNorm & w/o\\
        MNIST   & $93.8 \pm 0.69$ & $88.1 \pm 4.03$ & $98.8 \pm 0.08$ & $95.1 \pm 0.23$ & $99.1 \pm 0.08$ & $97.2 \pm 0.14 $ & $99.3 \pm 0.04$ & $98.9 \pm 0.04$ \\
        FMNIST  & $77.47 \pm 0.9$ & $74.9 \pm 2.07$ & $87.6 \pm 0.23$ & $76.3 \pm 3.03$ & $88.4 \pm 0.37$ & $80.1 \pm 2.95$ & $89.5 \pm 0.27$ & $77.9 \pm 12.26$ \\
        CIFAR10 & $53.5 \pm 0.76$ & $40.2 \pm 0.32$ & $63.6 \pm 0.39$ & $49.3 \pm 0.21$ & $68.8 \pm 0.32 $ & $56.3 \pm 3.14$ & $72.4 \pm 0.18$ & $64.3 \pm 1.08$ \\
        \bottomrule
    \end{tabular}
\end{table*}

\subsection{Energy savings}\label{sec:energyResults}
Energy cost in digital hardware can be divided into two categories: 
\emph{computational} and \emph{memory-access}.
Whether a given algorithm is compute- or memory-dominated depends on several architectural choices, such as 
dense layers vs.\ convolutional layers, skip connections, and batch normalization vs.\ instance normalization \cite{gholami2024ai,chenna2023evolution}.
In deep neural networks, computational energy is largely attributable to matrix multiplications, 
which involve multiply-and-add operations, which in turn is dominated by the cost of the multiplications alone \cite{horowitz20141}.
Thus, the \emph{number of multiplications} and the \emph{number of memory accesses} are the key factors determining energy costs. 

With binary stochastic forward-forward, we obtain energy savings over both backprop and real-valued forward-forward in both compute-dominated and memory-dominated regimes.
\tbl{energyCost} shows the dominant terms in both cases.
We justify these claims briefly here, deferring detailed derivations to \appref{energyConsumption}).
In the discussion below, $N$ is the batch size, $C$ the number of channels (approximated as equal across layers, $H, W$ the height and width of the input image, and $L$ the number of layers in the network.

\paragraph{Memory-dominated regime}
In the CwC architecture, standard backpropagation and (real-valued) CwC-FF consume roughly the same amount of memory.
This is because, unlike Hinton's FF, CwC-FF must still write to memory in the forward pass in order subsequently to compute the gradient through the BatchNorm operation.
This is true in our BSFF algorithm as well.
Nevertheless, BSFF saves roughly a factor of 32 by writing to memory binary, rather than 32-bit floating-point, numbers.
The first layer, operating on the real-valued inputs, does need to write 32-bit numbers, but it only occurs once (the $NC_\text{in}CHW$ term in row 3 of \tbl{energyCost}).
And since the number of input channels $C_\text{in}$ is between 1 and 3, this $NC_\text{in}CHW$ term will typically be about 50 to 100 times less than the $NC^2HW$ contribution of the non-input layers to the CwC-FF and backprop memory costs.
Hence the overall memory savings of BSFF are about a factor of 32.

\paragraph{Compute-dominated regime}
Multiplication of binary matrices amounts to indexing operations, which have negligible energy costs compared to real-valued matrix multiplications.
Therefore, in reckoning the total compute cost of BSFF, it is necessary only to identify the real-valued multiplications that cannot be eliminated.

All three algorithms incur a dominant cost of order \(O(NC^2K^2HW)\), resulting from the matrix multiplications in the convolutions.
In CwC-FF, each layer requires \(2NC^2K^2HW\) multiplications---one full convolution less than backpropagation, because the forward–forward algorithm does not propagate upstream gradients.
BSFF incurs this same cost, but \emph{only at the first layer}, where the inputs (image pixels) are 32-bit numbers (hence the omitted factor of $L$ from the first term in the table entry for BSFF multiplication costs).
After this, multiplications can be replaced by indexing operations, which have comparatively negligible costs.
Na{\"i}vely, the BatchNorm at each layer would reintroduce floating-point values.
However, we defer BatchNorm’s scale and shift until the next layer by absorbing them into its convolutional filters, at a cost of  $CK^2$ per channel (see Algorithm~\ref{alg:compute-gradient-BSN}, lines~\ref{alg:compute-gradient-BSN:dG-da} and~\ref{alg:compute-gradient-BSN:compute-dL_dW}).
This is the second term in the table entry for BSFF multiplication costs.

The first and second terms for BSFF multiplication costs are the same up to factors of $C_\text{in}HW$ and $2C(L-1)$, respectively.
For moderately deep networks ($L \approx 10)$ and modestly sized images (e.g.\ $32 \times 32$), these terms are of the same order.
Comparing the first term, then, with the multiplication cost for CwC-FF, we find a savings of $C/C_\text{in}L$, or approximately 1-2 orders of magnitude.

\paragraph{The cost of tiled logistic units}
All analysis of energy costs to this point assumes a single logistic unit per activation.
We now consider the effect of moving to ``tiled logistic units'' consisting of $M$ p-bits per activation (\sctn{approximatingReLUs}).
Evidently, memory costs scale as $\log_2 M$, the number of unique numbers representable by these units.
Thus in the most expensive case considered below, with $M=7$, memory costs only increase by a factor of 3.
This cuts the overall memory savings to about one order of magnitude.

Compute scaling is even more favorable.
Moving from logistic functions to $M$ tiled logistic functions introduces $M$ extra multiplications for every multiplication in the convolution.
But each new multiplication is between only a $\log_2 M$-bit number and a 32-bit number, whereas the original multiplication is between two 32-bit numbers.
The latter dominates for $M$ less than about 7 (see \appref{energyConsumption}: \nameref{ss:M-tiled-units}), so the tiled units introduce only negligible extra computational cost.
This leaves the total energy savings in the compute-dominated regime at about 1-2 orders of magnitude.

\subsection{Classification performance}
Here we compare the original real-valued, deterministic CwC-FF algorithm \cite{papachristodoulou2024convolutional} against our two binary stochastic variants:\ BSFF (\eqn{BSFFvariationalGradient})
and BGBSFF (\eqn{BGBSFFgradient}).
We consider both logistic units (BSFF:1, BGBSFF:1) and tiled logistic units, with either 2 (BSFF:2, BGBSFF:2), 3 (BSFF:3, BGBSFF:3), or 7 (BSFF:7, BGBSFF:7) tiled units (see \sctn{approximatingReLUs}).

For each variant of the algorithm, we train five different models from scratch.
\fig{accuracies} shows the results as boxplots.
We observe that, with a sufficient number of samples per tiled logistic unit, performance is roughly equal to performance with the standard (real-valued, deterministic) CwC-FF on all three datasets.
In the case of MNIST, two or three samples is sufficient to reach 99\% accuracy, although more are required to match ReLU performance for FMNIST and CIFAR-10.
Crucially, however, even BSFF:7 still achieves energy savings over CwC-FF of about 10$\times$ (if memory-bound) or 10--100$\times$ (if compute-bound).
So comparable performance to CwC-FF can be purchased at much lower energy costs.

We also note that using a binary stochastic gradient (BGBSFF) rather than a real-valued one (BSFF) does not much hurt performance on MNIST or FMNIST, for any number of tied binary-stochastic units.
On CIFAR-10, on the other hand, BSFF does consistently outperform BGBSFF (except for at BSFF-1).
But the additional energy cost of using BSFF instead of BGBSFF is small: the activation function’s Jacobian is diagonal, resulting in only $NCHW$ multiplications which is which is insignificant compared to the dominant term given in table \ref{tbl:energyCost}.
So the BGBSFF adjustment does not affect the dominant term in the total cost.

\subsection{Fully binary gradients}
On the other hand, if the BatchNorm is removed, BGBSFF achieves an order‑of‑magnitude energy saving, since all convolutional multiplications (including those in the first layer) are replaced by simple indexing operations—effectively eliminating the term in Table~\ref{tbl:energyCost} that accounts for the number of multiplications (see \appref{energyConsumption}: \nameref{ss:no-batchnorm}).
Moreover, without BatchNorm, each sample can be processed independently, so there is no need to re‑read the input for every output channel. Consequently, the dominant storage term reduces from $NC^2HW$ to $NCHW$, as detailed in \appref{energyConsumption}: \nameref{ss:no-batchnorm}.

Therefore, we investigate the effect of computing the CwC-FF loss at the max-pooling layer instead of at the batch normalization layer.
In this modification, we still z-score the outputs of the max pooling layer before passing them to the next layer, motivated by Hinton's original line of reasoning for forward-forward:\ if subsequent layers are to learn new features, the relative magnitudes of the classes must be equalized, but their directions retained.

The results of this experiment are shown in \tbl{lossAtMaxpool}.
With a sufficient number of samples (BGBSFF:7), performance on MNIST is comparable when applying the loss with or without BatchNorm.
But performance lags for the other two (more difficult) datasets; and the standard deviation across runs is generally higher.
Still, this result should be interpreted with caution, as no hyperparameter tuning was performed (see \sctn{details}).
We consider this an important avenue for future investigation.

\section{Discussion}\label{sec:discussion}

For decades, Moore’s Law drove computational gains through transistor miniaturization, with hardware and algorithm development evolving independently.
In the post-Moore era, closer integration of system design across all levels from physics to algorithms has become essential.

One promising direction in this integration is the \emph{p-bit}, a physical device that generates random Bernoulli samples at extremely high speeds and low energy costs.
This study has explored its potential for reducing the energy costs of discriminative classifiers, in particular by examining the performance of the forward-forward algorithm under stochastic binarization, which could in theory be implemented cheaply with p-bits.
Our binary-stochastic forward-forward algorithm reaches the same classification accuracy as a state-of-the-art forward-forward algorithm on simple datasets, at an estimated 1/10th to 1/100th the energy costs.
We see these results, which were generated purely in software, as critical for justifying and motivating the investment of a hardware implementation.

\subsection{Appraising the classification performance}
There are two important questions about accuracy that are not quite orthogonal:\ (1) how close is forward-forward (FF) to backprop? and (2) how close is binary stochastic FF to (deterministic) FF?
The main point of this study is to answer the second question (the answer is ``quite close''), but this second question would be uninteresting if the answer to the first question were ``not very close.''
For MNIST, the gap between FF and backprop is negligible; for CIFAR-10, it is about 10--20 percentage points (with the larger gap for very complex architectures).

For deterministic CwC-FF, this loss of accuracy would not be worth the very small gain in energy efficiency (recall \tbl{energyCost}).
But for 10--100$\times$ savings provided by BSFF, the price may be worth paying.
Whether that trade-off is worthwhile in any individual case depends on the relative costs of accuracy and energy.

Furthermore, there is reason to believe that FF algorithms will continue to close this gap, with our binary-stochastic variant close behind.
To our knowledge, no one has yet successfully extended forward-forward algorithms to very deep networks or datasets more complex than CIFAR-10.
But performance has been steadily improving since Hinton introduced FF in 2022.
Whereas that version achieved around 60\% accuracy on CIFAR‑10 and only about 6\% on CIFAR‑100, last year's CwC-FF \cite{papachristodoulou2024convolutional} (the point of departure for our study) achieved about 73\% and 50\%, resp.
Moreover, our technique for making FF binary and stochastic only negligibly reduces the accuracy of Hinton's FF, just as we have shown here that it only negligibly reduces the accuracy of CwC-FF.
Thus there is strong reason to believe that the technique works generally on FF algorithms, including future ones.
(We have not included the results for binary-stochasticizing Hinton's FF, since the overall results are worse than CwC-FF.)
There are also natural directions for improvement of FF, especially the integration of recurrent structures or of local predictive coding mechanisms, to enhance error minimization in deeper networks.
We are currently pursuing these.

\subsection{Memory vs.\ compute costs}
We have throughout remained agnostic about whether memory or compute will dominate energy costs in a hardware implementation of our algorithm, and accordingly have calculated both (\tbl{energyCost}, \appref{energyConsumption}).
However, this does matter since it determines whether the energy savings over backprop are, respectively, one or two orders of magnitude.
A task is \emph{compute‐dominated} if it performs many multiply–accumulate (MAC) operations per memory access: for example based on 65-nm CMOS measurements by Yang \emph{et al.}\cite{yang2017method}, one DRAM access costs roughly the same energy as \(\approx200\) MACs, whereas one SRAM access costs \(\approx6\) MACs. Accordingly, when each DRAM (resp.\ SRAM) access is amortized over more than 200 (resp.\ 6) MACs, compute energy exceeds memory energy. In shallow CNNs (up to 4–5 layers), weights, activations, gradients, and other intermediate tensors typically fit on‑chip, making off‑chip DRAM traffic negligible and the networks compute‐dominated. In mid‑size networks (6–10 layers), on‑chip capacity is insufficient to store all activations required for backpropagation, increasing DRAM accesses such that memory energy becomes comparable to compute energy. In large networks $> 10$ layers), frequent off‑chip transfers of data make these networks memory‐dominated. Thus, depending on network depth and the computation‑to‑memory ratio, either compute or memory may dominate the total energy cost, a trade‑off that underlies our evaluation of energy savings relative to conventional backpropagation.

\subsection{Analog hardware}
We have implicitly assumed an implementation on a digital architecture, but even greater energy savings are theoretically achievable in more exotic architectures.
For example, in a multi-chip setup where each layer resides on a separate chip, backpropagation introduces sequential dependencies, requiring the first-layer chip to wait for results from the last layer.
This increases memory bandwidth demands and access costs.
In contrast, forward-forward, with its local updates, allows for parallel execution across layers, significantly reducing the need for distant memory accesses.
Analog multiplications can also significantly increase energy efficiency.
Currently, operations like batch normalization require large memory readouts, limiting efficiency in purely analog systems.
One promising research direction is to develop an \emph{analog-friendly} substitute for batch normalization, allowing outputs to be processed locally before being passed to the next layer.
This could yield even greater cost reductions.

More broadly, the pursuit of hardware-friendly learning algorithms presents a significant opportunity for future research.
Achieving truly energy-efficient computation will require interdisciplinary expertise spanning hardware design, alternative learning paradigms, and circuit optimization.
By bridging these fields, we can unlock substantial gains in energy efficiency and computational scalability, addressing the demands of modern day AI needs.

\section*{Conflict of Interest}
[Redacted for anonymity] has a financial interest in [Redacted].


%



\ifCLASSOPTIONcaptionsoff
  \newpage
\fi



%

\bibliographystyle{IEEEtran}
\bibliography{%
    \bibsdir/articles,%
    \bibsdir/MachineLearning_acceleration,%
    \bibsdir/ML%
}

\begin{thebibliography}{10}
\providecommand{\url}[1]{#1}
\csname url@samestyle\endcsname
\providecommand{\newblock}{\relax}
\providecommand{\bibinfo}[2]{#2}
\providecommand{\BIBentrySTDinterwordspacing}{\spaceskip=0pt\relax}
\providecommand{\BIBentryALTinterwordstretchfactor}{4}
\providecommand{\BIBentryALTinterwordspacing}{\spaceskip=\fontdimen2\font plus
\BIBentryALTinterwordstretchfactor\fontdimen3\font minus \fontdimen4\font\relax}
\providecommand{\BIBforeignlanguage}[2]{{%
\expandafter\ifx\csname l@#1\endcsname\relax
\typeout{** WARNING: IEEEtran.bst: No hyphenation pattern has been}%
\typeout{** loaded for the language `#1'. Using the pattern for}%
\typeout{** the default language instead.}%
\else
\language=\csname l@#1\endcsname
\fi
#2}}
\providecommand{\BIBdecl}{\relax}
\BIBdecl

\bibitem{Devlin2018}
\BIBentryALTinterwordspacing
J.~Devlin, M.-W. Chang, K.~Lee, and K.~Toutanova, ``Bert: Pre-training of deep bidirectional transformers for language understanding,'' 10 2018. [Online]. Available: \url{http://arxiv.org/abs/1810.04805}
\BIBentrySTDinterwordspacing

\bibitem{krizhevsky2012imagenet}
A.~Krizhevsky, I.~Sutskever, and G.~E. Hinton, ``Imagenet classification with deep convolutional neural networks,'' \emph{Advances in neural information processing systems}, vol.~25, 2012.

\bibitem{Hannun2014}
\BIBentryALTinterwordspacing
A.~Hannun, C.~Case, J.~Casper, B.~Catanzaro, G.~Diamos, E.~Elsen, R.~Prenger, S.~Satheesh, S.~Sengupta, A.~Coates, and A.~Y. Ng, ``Deep speech: Scaling up end-to-end speech recognition,'' 12 2014. [Online]. Available: \url{http://arxiv.org/abs/1412.5567}
\BIBentrySTDinterwordspacing

\bibitem{Jumper2021}
J.~Jumper, R.~Evans, A.~Pritzel, T.~Green, M.~Figurnov, O.~Ronneberger, K.~Tunyasuvunakool, R.~Bates, A.~Žídek, A.~Potapenko, A.~Bridgland, C.~Meyer, S.~A. Kohl, A.~J. Ballard, A.~Cowie, B.~Romera-Paredes, S.~Nikolov, R.~Jain, J.~Adler, T.~Back, S.~Petersen, D.~Reiman, E.~Clancy, M.~Zielinski, M.~Steinegger, M.~Pacholska, T.~Berghammer, S.~Bodenstein, D.~Silver, O.~Vinyals, A.~W. Senior, K.~Kavukcuoglu, P.~Kohli, and D.~Hassabis, ``Highly accurate protein structure prediction with alphafold,'' \emph{Nature}, vol. 596, pp. 583--589, 8 2021.

\bibitem{economist2020cost}
``The cost of training machines is becoming a problem,'' \emph{{The Economist}}, 2020, accessed: 2024-08-12.

\bibitem{motherjones2024energy}
D.~Milmo, A.~Hern, and J.~Ambrose, ``Ai's growing demand for data centers is threatening tech's climate goals,'' \href{https://www.motherjones.com/environment/2024/07/ai-data-centers-energy-use-tech-climate-goals-net-zero/}, Jul. 2024, accessed: 2024-08-12.

\bibitem{MIT2024energy}
C.~Crownhart, ``Ai is an energy hog. this is what it means for climate change,'' \href{https://www.technologyreview.com/2024/05/23/1092777/ai-is-an-energy-hog-this-is-what-it-means-for-climate-change/}, May 2024, accessed: 2024-08-12.

\bibitem{Howard2017}
\BIBentryALTinterwordspacing
A.~G. Howard, M.~Zhu, B.~Chen, D.~Kalenichenko, W.~Wang, T.~Weyand, M.~Andreetto, and H.~Adam, ``Mobilenets: Efficient convolutional neural networks for mobile vision applications,'' 4 2017. [Online]. Available: \url{http://arxiv.org/abs/1704.04861}
\BIBentrySTDinterwordspacing

\bibitem{li2019fully}
R.~Li, Y.~Wang, F.~Liang, H.~Qin, J.~Yan, and R.~Fan, ``Fully quantized network for object detection,'' in \emph{Proceedings of the IEEE/CVF conference on computer vision and pattern recognition}, 2019, pp. 2810--2819.

\bibitem{Ankit2019}
A.~Ankit, I.~E. Hajj, S.~R. Chalamalasetti, G.~Ndu, M.~Foltin, R.~S. Williams, P.~Faraboschi, W.~M. Hwu, J.~P. Strachan, K.~Roy, and D.~S. Milojicic, ``Puma: A programmable ultra-efficient memristor-based accelerator for machine learning inference,'' in \emph{International Conference on Architectural Support for Programming Languages and Operating Systems - ASPLOS}.\hskip 1em plus 0.5em minus 0.4em\relax Association for Computing Machinery, 4 2019, pp. 715--731.

\bibitem{gholami2022survey}
A.~Gholami, S.~Kim, Z.~Dong, Z.~Yao, M.~W. Mahoney, and K.~Keutzer, ``A survey of quantization methods for efficient neural network inference,'' in \emph{Low-Power Computer Vision}.\hskip 1em plus 0.5em minus 0.4em\relax Chapman and Hall/CRC, 2022, pp. 291--326.

\bibitem{Hubara2018}
I.~Hubara, M.~Courbariaux, D.~Soudry, R.~El-Yaniv, and Y.~Bengio, ``Quantized neural networks: Training neural networks with low precision weights and activations,'' pp. 1--30, 2018.

\bibitem{Rumelhart1986}
D.~E. Rumelhart, G.~E. Hinton, and R.~J. Williams, ``Learning representations by back-propagating errors,'' \emph{nature}, vol. 323, no. 6088, pp. 533--536, 1986.

\bibitem{Aguirre2024}
F.~Aguirre, A.~Sebastian, M.~L. Gallo, W.~Song, T.~Wang, J.~J. Yang, W.~Lu, M.~F. Chang, D.~Ielmini, Y.~Yang, A.~Mehonic, A.~Kenyon, M.~A. Villena, J.~B. Roldán, Y.~Wu, H.~H. Hsu, N.~Raghavan, J.~Suñé, E.~Miranda, A.~Eltawil, G.~Setti, K.~Smagulova, K.~N. Salama, O.~Krestinskaya, X.~Yan, K.~W. Ang, S.~Jain, S.~Li, O.~Alharbi, S.~Pazos, and M.~Lanza, ``Hardware implementation of memristor-based artificial neural networks,'' 12 2024.

\bibitem{Tsai2018}
H.~Tsai, S.~Ambrogio, P.~Narayanan, R.~M. Shelby, and G.~W. Burr, ``Recent progress in analog memory-based accelerators for deep learning,'' 6 2018.

\bibitem{Hu2016}
M.~Hu, J.~P. Strachan, Z.~Li, E.~M. Grafals, N.~Davila, C.~Graves, S.~Lam, N.~Ge, J.~J. Yang, and R.~S. Williams, ``Dot-product engine for neuromorphic computing: Programming 1t1m crossbar to accelerate matrix-vector multiplication,'' in \emph{Proceedings - Design Automation Conference}, vol. 05-09-June-2016.\hskip 1em plus 0.5em minus 0.4em\relax Institute of Electrical and Electronics Engineers Inc., 6 2016.

\bibitem{Shafiee2016}
A.~Shafiee, A.~Nag, N.~Muralimanohar, R.~Balasubramonian, J.~P. Strachan, M.~Hu, R.~S. Williams, and V.~Srikumar, ``Isaac: A convolutional neural network accelerator with in-situ analog arithmetic in crossbars,'' in \emph{Proceedings - 2016 43rd International Symposium on Computer Architecture, ISCA 2016}.\hskip 1em plus 0.5em minus 0.4em\relax Institute of Electrical and Electronics Engineers Inc., 8 2016, pp. 14--26.

\bibitem{Frenkel2021}
C.~Frenkel, M.~Lefebvre, and D.~Bol, ``Learning without feedback: Fixed random learning signals allow for feedforward training of deep neural networks,'' \emph{Frontiers in Neuroscience}, vol.~15, 2 2021.

\bibitem{Hinton2022}
\BIBentryALTinterwordspacing
G.~Hinton, ``The forward-forward algorithm: Some preliminary investigations,'' 12 2022. [Online]. Available: \url{http://arxiv.org/abs/2212.13345}
\BIBentrySTDinterwordspacing

\bibitem{Kohan2022}
\BIBentryALTinterwordspacing
A.~Kohan, E.~A. Rietman, and H.~T. Siegelmann, ``Signal propagation: A framework for learning and inference in a forward pass,'' 4 2022. [Online]. Available: \url{http://arxiv.org/abs/2204.01723}
\BIBentrySTDinterwordspacing

\bibitem{Dellaferrera2022}
\BIBentryALTinterwordspacing
G.~Dellaferrera and G.~Kreiman, ``Error-driven input modulation: Solving the credit assignment problem without a backward pass,'' 1 2022. [Online]. Available: \url{http://arxiv.org/abs/2201.11665}
\BIBentrySTDinterwordspacing

\bibitem{jiang2023one}
J.~Jiang, Z.~Zhang, C.~Xu, Z.~Yu, and Y.~Peng, ``One forward is enough for neural network training via likelihood ratio method,'' \emph{arXiv preprint arXiv:2305.08960}, 2023.

\bibitem{Chatterjee2019}
B.~Chatterjee, P.~Panda, S.~Maity, A.~Biswas, K.~Roy, and S.~Sen, ``Exploiting inherent error resiliency of deep neural networks to achieve extreme energy efficiency through mixed-signal neurons,'' \emph{IEEE Transactions on Very Large Scale Integration (VLSI) Systems}, vol.~27, no.~6, pp. 1365--1377, 2019.

\bibitem{Courbariaux2014}
\BIBentryALTinterwordspacing
M.~Courbariaux, Y.~Bengio, and J.-P. David, ``Training deep neural networks with low precision multiplications,'' 12 2014. [Online]. Available: \url{http://arxiv.org/abs/1412.7024}
\BIBentrySTDinterwordspacing

\bibitem{Qin2020}
\BIBentryALTinterwordspacing
H.~Qin, R.~Gong, X.~Liu, X.~Bai, J.~Song, and N.~Sebe, ``Binary neural networks: A survey,'' 3 2020. [Online]. Available: \url{http://arxiv.org/abs/2004.03333 http://dx.doi.org/10.1016/j.patcog.2020.107281}
\BIBentrySTDinterwordspacing

\bibitem{Peters2018}
\BIBentryALTinterwordspacing
J.~W.~T. Peters and M.~Welling, ``Probabilistic binary neural networks,'' 9 2018. [Online]. Available: \url{http://arxiv.org/abs/1809.03368}
\BIBentrySTDinterwordspacing

\bibitem{zhao2017accelerating}
R.~Zhao, W.~Song, W.~Zhang, T.~Xing, J.-H. Lin, M.~Srivastava, R.~Gupta, and Z.~Zhang, ``Accelerating binarized convolutional neural networks with software-programmable fpgas,'' in \emph{Proceedings of the 2017 ACM/SIGDA international symposium on field-programmable gate arrays}, 2017, pp. 15--24.

\bibitem{nurvitadhi2016accelerating}
E.~Nurvitadhi, D.~Sheffield, J.~Sim, A.~Mishra, G.~Venkatesh, and D.~Marr, ``Accelerating binarized neural networks: Comparison of fpga, cpu, gpu, and asic,'' in \emph{2016 International Conference on Field-Programmable Technology (FPT)}.\hskip 1em plus 0.5em minus 0.4em\relax IEEE, 2016, pp. 77--84.

\bibitem{zhou2017deep}
Y.~Zhou, S.~Redkar, and X.~Huang, ``Deep learning binary neural network on an fpga,'' in \emph{2017 IEEE 60th International Midwest Symposium on Circuits and Systems (MWSCAS)}.\hskip 1em plus 0.5em minus 0.4em\relax IEEE, 2017, pp. 281--284.

\bibitem{Kaiser2019}
J.~Kaiser, A.~Rustagi, K.~Y. Camsari, J.~Z. Sun, S.~Datta, and P.~Upadhyaya, ``Subnanosecond fluctuations in low-barrier nanomagnets,'' \emph{Physical Review Applied}, vol.~12, 11 2019.

\bibitem{papachristodoulou2024convolutional}
A.~Papachristodoulou, C.~Kyrkou, S.~Timotheou, and T.~Theocharides, ``Convolutional channel-wise competitive learning for the forward-forward algorithm,'' in \emph{Proceedings of the AAAI Conference on Artificial Intelligence}, vol.~38, no.~13, 2024, pp. 14\,536--14\,544.

\bibitem{Lillicrap2016}
T.~P. Lillicrap, D.~Cownden, D.~B. Tweed, and C.~J. Akerman, ``Random synaptic feedback weights support error backpropagation for deep learning,'' \emph{Nature Communications}, vol.~7, 11 2016.

\bibitem{Nokland2016}
\BIBentryALTinterwordspacing
A.~Nøkland, ``Direct feedback alignment provides learning in deep neural networks,'' 9 2016. [Online]. Available: \url{http://arxiv.org/abs/1609.01596}
\BIBentrySTDinterwordspacing

\bibitem{zhao2023cascaded}
G.~Zhao, T.~Wang, Y.~Li, Y.~Jin, C.~Lang, and H.~Ling, ``The cascaded forward algorithm for neural network training,'' \emph{arXiv preprint arXiv:2303.09728}, 2023.

\bibitem{Scellier2017}
B.~Scellier and Y.~Bengio, ``Equilibrium propagation: Bridging the gap between energy-based models and backpropagation,'' \emph{Frontiers in Computational Neuroscience}, vol.~11, 5 2017.

\bibitem{Scellier2022}
\BIBentryALTinterwordspacing
B.~Scellier, S.~Mishra, Y.~Bengio, and Y.~Ollivier, ``Agnostic physics-driven deep learning,'' 5 2022. [Online]. Available: \url{http://arxiv.org/abs/2205.15021}
\BIBentrySTDinterwordspacing

\bibitem{Bengio2013}
\BIBentryALTinterwordspacing
Y.~Bengio, N.~Léonard, and A.~Courville, ``Estimating or propagating gradients through stochastic neurons for conditional computation,'' 8 2013. [Online]. Available: \url{http://arxiv.org/abs/1308.3432}
\BIBentrySTDinterwordspacing

\bibitem{Shekhovtsov2020}
\BIBentryALTinterwordspacing
A.~Shekhovtsov and V.~Yanush, ``Reintroducing straight-through estimators as principled methods for stochastic binary networks,'' 6 2020. [Online]. Available: \url{http://arxiv.org/abs/2006.06880}
\BIBentrySTDinterwordspacing

\bibitem{Sutton2020}
B.~Sutton, R.~Faria, L.~A. Ghantasala, R.~Jaiswal, K.~Y. Camsari, and S.~Datta, ``Autonomous probabilistic coprocessing with petaflips per second,'' \emph{IEEE Access}, vol.~8, pp. 157\,238--157\,252, 2020.

\bibitem{kaiser2021probabilistic}
J.~Kaiser and S.~Datta, ``Probabilistic computing with p-bits,'' \emph{Applied Physics Letters}, vol. 119, no.~15, 2021.

\bibitem{Aadit2021}
\BIBentryALTinterwordspacing
N.~A. Aadit, A.~Grimaldi, M.~Carpentieri, L.~Theogarajan, J.~M. Martinis, G.~Finocchio, and K.~Y. Camsari, ``Massively parallel probabilistic computing with sparse ising machines,'' 10 2021. [Online]. Available: \url{http://arxiv.org/abs/2110.02481 http://dx.doi.org/10.1038/s41928-022-00774-2}
\BIBentrySTDinterwordspacing

\bibitem{Camsari2017}
K.~Y. Camsari, R.~Faria, B.~M. Sutton, and S.~Datta, ``Stochastic p-bits for invertible logic,'' \emph{Physical Review X}, vol.~7, 7 2017.

\bibitem{Borders2019}
W.~A. Borders, A.~Z. Pervaiz, S.~Fukami, K.~Y. Camsari, H.~Ohno, and S.~Datta, ``Integer factorization using stochastic magnetic tunnel junctions,'' \emph{Nature}, vol. 573, pp. 390--393, 9 2019.

\bibitem{Neal1998}
R.~M. Neal and G.~E. Hinton, ``{A view of the EM algorithm that justifies incremental, sparse, and other variants},'' \emph{Learning in Graphical Models}, 1998.

\bibitem{Ioffe2015}
\BIBentryALTinterwordspacing
S.~Ioffe and C.~Szegedy, ``Batch normalization: Accelerating deep network training by reducing internal covariate shift,'' 2015. [Online]. Available: \url{https://arxiv.org/abs/1502.03167}
\BIBentrySTDinterwordspacing

\bibitem{deng2012mnist}
L.~Deng, ``The mnist database of handwritten digit images for machine learning research,'' \emph{IEEE Signal Processing Magazine}, vol.~29, no.~6, pp. 141--142, 2012.

\bibitem{xiao2017fashionmnist}
\BIBentryALTinterwordspacing
H.~Xiao, K.~Rasul, and R.~Vollgraf, ``Fashion-mnist: a novel image dataset for benchmarking machine learning algorithms,'' 2017. [Online]. Available: \url{https://github.com/zalandoresearch/fashion-mnist}
\BIBentrySTDinterwordspacing

\bibitem{krizhevsky2009learning}
A.~Krizhevsky, G.~Hinton \emph{et~al.}, ``Learning multiple layers of features from tiny images,'' 2009.

\bibitem{gholami2024ai}
A.~Gholami, Z.~Yao, S.~Kim, C.~Hooper, M.~W. Mahoney, and K.~Keutzer, ``Ai and memory wall,'' \emph{IEEE Micro}, 2024.

\bibitem{chenna2023evolution}
D.~Chenna, ``Evolution of convolutional neural network (cnn): Compute vs memory bandwidth for edge ai,'' \emph{arXiv preprint arXiv:2311.12816}, 2023.

\bibitem{horowitz20141}
M.~Horowitz, ``1.1 computing's energy problem (and what we can do about it),'' in \emph{2014 IEEE international solid-state circuits conference digest of technical papers (ISSCC)}.\hskip 1em plus 0.5em minus 0.4em\relax IEEE, 2014, pp. 10--14.

\bibitem{yang2017method}
T.-J. Yang, Y.-H. Chen, J.~Emer, and V.~Sze, ``A method to estimate the energy consumption of deep neural networks,'' in \emph{2017 51st asilomar conference on signals, systems, and computers}.\hskip 1em plus 0.5em minus 0.4em\relax IEEE, 2017, pp. 1916--1920.

\end{thebibliography}

\appendices
\onecolumn


\section{Loss Functions for Binary Stochastic Forward-Forward}\label{sec:losses}
As discussed in the main text, introducing stochastic variables $\Generltnts$ into the discriminative model requires marginalizing them back out in order to compute the loss, \eqn{marginalRelativeEntropy}, or its gradient.
The loss itself can be written as
\begin{equation}\label{eqn:BSFFgradient}
    \begin{split}
        \colttlderiv{\mathcal{L}}{\inwts{l}}
            &=
        \def\integrand#1 {
            \frac{1}{\genermarginal{} }
            \colttlderiv{}{\inwts{l}}
            \genermarginal{#1}
        }
        -\sampleaverage{patent/\Dataobsvs,conditioner/\Dataconditioners}{\integrand}\\
            &=
        \def\integranda#1 {
            \generemission{#1}
            \colttlderiv{}{\inwts{l}}
            \generprior{#1}
        }        
        \def\integrandb#1 {
            \frac{1}{\genermarginal{#1} }
            \dmarginalize{latent/\generltnts}{\integranda#1,}
        }
        -\sampleaverage{patent/\Dataobsvs,conditioner/\Dataconditioners}{\integrandb}\\
            &=
        \def\integranda#1 {
            \generemission{#1}
            \generprior{#1}
            \colttlderiv{}{\inwts{l}}
            \log\generpriorfactor{#1,latent=\generltnt{l}}
        }        
        \def\integrandb#1 {
            \frac{1}{\genermarginal{#1} }
            \dmarginalize{latent/\generltnts}{\integranda#1,}
        }
        -\sampleaverage{patent/\Dataobsvs,conditioner/\Dataconditioners}{\integrandb}\\
            &=
        \def\integranda#1 {
            \generposterior{#1}
            \colttlderiv{}{\inwts{l}}
            \log\generpriorfactor{#1,latent=\generltnt{l}}
        }        
        \def\integrandb#1 {
            \dmarginalize{latent/\generltnts}{\integranda#1,}
        }
        -\sampleaverage{patent/\Dataobsvs,conditioner/\Dataconditioners}{\integrandb},
    \end{split}
\end{equation}
The expectation under $\generposterior{} $ could be approximated by a sample average, except that computing the normalizer for this distribution is computationally intractable.
Instead we can interpret the penultimate line as an importance sampler, with $\generprior{} $ as the proposal distribution and $\generemission{} $
as the importance weight.
This is tractable because $\generprior{} $ factorizes (\eqn{independentBernoullis}), so samples can be drawn independently.
In terms of the importance weight, the normalizer can be written as
\begin{equation}\label{eqn:importanceSampledNormalizer}
    \begin{split}
        \genermarginal{}
            =
        \def\integranda#1 {\generemission{#1} \generprior{#1}} 
        \dmarginalize{binary/\generltnts}{\integranda}.
    \end{split}
\end{equation}
Putting \eqns{BSFFgradient}{importanceSampledNormalizer} together with an explicit expression for
$\colttlderivflat{(\log\generpriorfactor{index=l} )}{\inwts{l}} $ yields
\begin{equation*}
    \colttlderiv{\mathcal{L}}{\inwts{l}}
        =
    \def\integranda#1 {
        \assignkeys{distributions, gener, adjust, #1}%
        \generemission{#1}
        \left(\real{l} - \Generltnt{l}\right)\conditioner
    }
    \def\integrandc#1 {
        \assignkeys{distributions, gener, adjust, #1}%
        \frac{%
            \condsampleaverage{latent/\Generltnts}{ppatent/\conditioner}{\integranda#1,}
        }{%
            \condsampleaverage{latent/\Generltnts}{ppatent/\conditioner}{\generemission#1,}        
        }
    }
    \sampleaverage{patent/\Dataobsvs,conditioner/\Dataconditioners}{\integrandc} .
\end{equation*}

In our experiments, however, we did not achieve competitive results with this importance sampler.
Alternatively, then, we can attempt to minimize the following upper bound:
\begin{equation}\label{eqn:jointRelativeEntropyLong}
    \begin{split}
        \mathcal{J} 
            &\defeqleft
        \def\integranda#1 {\datamarginal{#1} \generprior{#1} }
        \relativeentropy{latent/\Generltnts,conditioner/\Dataconditioners,patent=\Dataobsvs}{\integranda}{\generjoint}\\
            &=
        \def\integranda#1 {\datamarginal{#1} \generprior{#1} }
        \def\integrandb#1 {\generemission{#1} \generprior{#1} }
        \relativeentropy{latent/\Generltnts,conditioner/\Dataconditioners,patent=\Dataobsvs}{\integranda}{\integrandb}\\
            &=
        \relativeentropy{latent/\Generltnts,conditioner/\Dataconditioners,patent=\Dataobsvs}{\datamarginal}{\generemission}\\
            &=
        \def\integranda#1 {
            \generprior{#1,paramdisplay={}}
            \log\generemission{#1,paramdisplay={}}
        }
        \def\integrandb#1 {
            \dmarginalize{latent/\generltnts}{\integranda #1,}
        }
        \sampleaverage{patent/\Dataobsvs,conditioner/\Dataconditioners}{\integrandb} + C.
    \end{split}
\end{equation}
This is \eqn{jointRelativeEntropy} in the main text.
It is an upper bound on the relative entropy in \eqn{marginalRelativeEntropy} because it can be expressed as the sum of that relative entropy and another relative entropy:
\begin{equation*}
    \begin{split}
        \mathcal{J}
            &=
        \def\integranda#1 {\datamarginal{#1} \generprior{#1} }
        \def\integrandb#1 {\genermarginal{#1} \generposterior{#1} }
        \relativeentropy{latent/\Generltnts,conditioner/\Dataconditioners,patent=\Dataobsvs}{\integranda}{\integrandb}\\
            &=
        \relativeentropy{latent/\Generltnts,conditioner/\Dataconditioners,patent=\Dataobsvs}{\datamarginal}{\genermarginal}
            +
        \relativeentropy{latent/\Generltnts,conditioner/\Dataconditioners,patent=\Dataobsvs}{\generprior}{\generposterior}.
    \end{split}
\end{equation*}

To compute the gradient of the joint relative entropy, \eqn{jointRelativeEntropyLong}, with respect to the $\lth$ row $\inwts{l}$ of the input matrix $\INWT$, we exploit the independence of the elements of $\generltnts$ and consider $\generltnt{l}$ separately from the remaining elements, ${\generltnts_{\backslash l}}$:
\begin{equation}\label{eqn:jointRelativeEntropyGradient}
    \begin{split}
        \colttlderiv{\mathcal{J}}{\inwts{l}}
            &=
        \def\integranda#1 {
            \generprior{#1,paramdisplay={}}
            \log\generemission{#1,paramdisplay={}}
        }
        \def\integrandb#1 {
            \dmarginalize{latent/\generltnts}{\integranda #1,}
        }
        -\colttlderiv{}{\inwts{l}}
        \sampleaverage{patent/\Dataobsvs,conditioner/\Dataconditioners}{\integrandb}\\
            &=
        \def\integranda#1 {
            \generpriorfactor{#1,latent=\onelatent,paramdisplay={}}
            \log\generemission{#1,latent={\LOOlatent,\onelatent},paramdisplay={}}
        }
        \def\integrandb#1 {
            \generprior{#1,latent=\LOOlatent,paramdisplay={}}
            \colttlderiv{}{\inwts{l}}
            \dmarginalize{onelatent/\generltnt{l}}{\integranda #1,}
        }
        \def\integrandc#1 {
            \dmarginalize{LOOlatent/{\generltnts_{\backslash l}}}{\integrandb #1,}
        }
        -\sampleaverage{patent/\Dataobsvs,conditioner/\Dataconditioners}{\integrandc}\\
            &=
        \def\integrandb#1 {
            \generprior{#1,latent=\LOOlatent,paramdisplay={}}
            \colttlderiv{}{\inwts{l}}
            \left(
                \real{l}\log\generemission{#1,latent={\LOOlatent,\Generltnt{l}=1},paramdisplay={}}
                    +
                (1 - \real{l})\log\generemission{#1,latent={\LOOlatent,\Generltnt{l}=0},paramdisplay={}}
            \right)
        }
        \def\integrandc#1 {
            \dmarginalize{LOOlatent/{\generltnts_{\backslash l}}}{\integrandb #1,}
        }
        -\sampleaverage{patent/\Dataobsvs,conditioner/\Dataconditioners}{\integrandc}\\
            &=
        \def\integrandb#1 {
            \assignkeys{distributions, gener, adjust, #1}%
            \generprior{#1,latent=\LOOlatent,paramdisplay={}}
            \left[
                -\log\generemission{#1,latent={\LOOlatent,\Generltnt{l}=1},paramdisplay={}}
                    +
                \log\generemission{#1,latent={\LOOlatent,\Generltnt{l}=0},paramdisplay={}}
            \right]
            \real{l}(1 - \real{l})\conditioner
        }
        \def\integrandc#1 {
            \dmarginalize{LOOlatent/{\generltnts_{\backslash l}}}{\integrandb #1,}
        }
        \sampleaverage{patent/\Dataobsvs,conditioner/\Dataconditioners}{\integrandc}.
    \end{split}
\end{equation}
The expectation can be carried out explicitly (third line) and differentiated (fourth line) because $\Generltnt{l}$ has only two possible values.

The bracketed quantity on the last line of \eqn{jointRelativeEntropyGradient} is the increase in surprisal when the $\lth$ element of $\generltnts$ is flipped from 0 to 1.
From the definition of the model output, \eqn{independentBernoullis}, the surprisal itself is
\begin{equation*}
    -\log\generemission{}
        =
    \logpartitionop{\OUTWT\argltnts} - \argobsvs\tr\OUTWT\argltnts - \log(h(\argobsvs)),
\end{equation*}
so the increase in surprisal is
\begin{equation}\label{eqn:deltaSurprisal}
    \logpartitionop{\sum_{\hiddim\neq l}^{\Hiddim}\outwts{d}\argltnt{d} + \outwts{l}} -
    \logpartitionop{\sum_{\hiddim\neq l}^{\Hiddim}\outwts{d}\argltnt{d}}
     -
    \argobsvs\tr\outwts{l}.
\end{equation}
Now recall from \eqns{HintonReadoutWeights}{GreekReadoutWeights} that the readout weights are inversely proportional to the hidden dimension $\Hiddim$.
Thus for large numbers of hidden units, the difference of log-partition functions in \eqn{deltaSurprisal} approaches the directional derivative.
To make this explicit, we introduce unscaled weights \barwts{d} such that $\outwts{d} = \barwts{d}/\Hiddim$.
Then as the size of the hidden dimension grows large, the difference becomes
\begin{equation*}
    \lim_{\Hiddim\to\infty} 
    \left[
    \frac{
        \logpartitionop{\sum_{\hiddim\neq l}^{\Hiddim}\outwts{d}\argltnt{d}
            + \frac{1}{\Hiddim}\barwts{l}} -
        \logpartitionop{\sum_{\hiddim\neq l}^{\Hiddim}\outwts{d}\argltnt{d}}
    }{1/\Hiddim}
    \frac{1}{\Hiddim}
    \right]
        =
    \colgradient{\logpartitionsym}{\ntrlparams{}} \cdot
    \barwts{l}
    \frac{1}{\Hiddim}
        =
    \colgradient{\logpartitionsym}{\ntrlparams{}} \cdot
    \outwts{l}.
\end{equation*}
Substituting this quantity into the difference of surprisals \eqn{deltaSurprisal} and thence into the gradient \eqn{jointRelativeEntropyGradient}, we obtain
\begin{equation}\label{eqn:BSFFvariationalGradientLong}
    \begin{split}
        \colttlderiv{\mathcal{J}}{\inwts{l}}
            &\approx
        \def\integrandb#1 {
            \assignkeys{distributions, gener, adjust, #1}%
            \generprior{#1,paramdisplay={}}
            \left(\colgradient{\logpartitionsym}{\ntrlparams} - \patent\right)\tr\outwts{l}
            \real{l}(1 - \real{l})\conditioner
        }
        \def\integrandc#1 {
            \dmarginalize{latent/{\generltnts_{\backslash l}}}{\integrandb #1,}
        }
        \sampleaverage{patent/\Dataobsvs,conditioner/\Dataconditioners}{\integrandc}\\
            &\approx
        \def\integrandb#1 {
            \assignkeys{distributions, gener, adjust, #1}%
            \generprior{#1,paramdisplay={}}
            \left(\colgradient{\logpartitionsym}{\ntrlparams} - \patent\right)\tr\outwts{l}
            \real{l}(1 - \real{l})\conditioner
        }
        \def\integrandc#1 {
            \dmarginalize{latent/\generltnts}{\integrandb #1,}
        }
        \sampleaverage{patent/\Dataobsvs,conditioner/\Dataconditioners}{\integrandc}\\
            &=
        \def\integranda#1 {%
            \assignkeys{distributions, gener, adjust, #1}%
            \patent
        }
        \def\integrandb#1 {%
            \assignkeys{distributions, gener, adjust, #1}%
            \condexpectation{patent/\Generobsvs}{\Generltnts}{\integranda #1,}{caca/\latent}
        }
        \def\integrandc#1 {
            \assignkeys{distributions, gener, adjust, #1}%
            \left(
                \condexpectation{latent/\Generltnts}{\conditioner}{\integrandb #1,}{foo/\conditioner}
            - \patent
            \right)\tr\outwts{l}
            \real{l}(1 - \real{l})\conditioner
        }
        \sampleaverage{patent/\Dataobsvs,conditioner/\Dataconditioners}{\integrandc}\\
            &=
        \def\integrandb#1 {%
            \assignkeys{distributions, gener, adjust, #1}%
            \inverselink{\OUTWT\latent}
        }
        \def\integrandc#1 {
            \assignkeys{distributions, gener, adjust, #1}%
            \left(
                \condexpectation{latent/\Generltnts}{\conditioner}{\integrandb #1,}{foo/\conditioner}
            - \patent
            \right)\tr\outwts{l}
            \real{l}(1 - \real{l})\conditioner
        }
        \sampleaverage{patent/\Dataobsvs,conditioner/\Dataconditioners}{\integrandc}.
    \end{split}
\end{equation}
The second line follows because, for a large number of hidden units, omitting or retaining a single row $\barwts{l}$ does not much affect the gradient of the log-partition function.
The final lines follow, as in \eqn{LFFgradient}, by substituting either the expectation of the sufficient statistics (penultimate line), or the inverse link ($\inverselink{\cdot}$) applied to the natural parameters (final line), for the gradient of the log-partition function.
This is \eqn{BSFFvariationalGradient} in the main text.

\newpage
\section{Energy consumption:\ a comparison of multiplication and memory operations}\label{sec:energyConsumption}
\textbf{Notation Disclaimer*}
The notation used in this appendix is defined within the local context of this section and may differ from or be independent of the notation used in the main text.
Readers should interpret symbols and definitions as specific to this appendix to avoid any confusion.

\begin{figure}[t]
  \centering
  \includegraphics[width=\columnwidth]{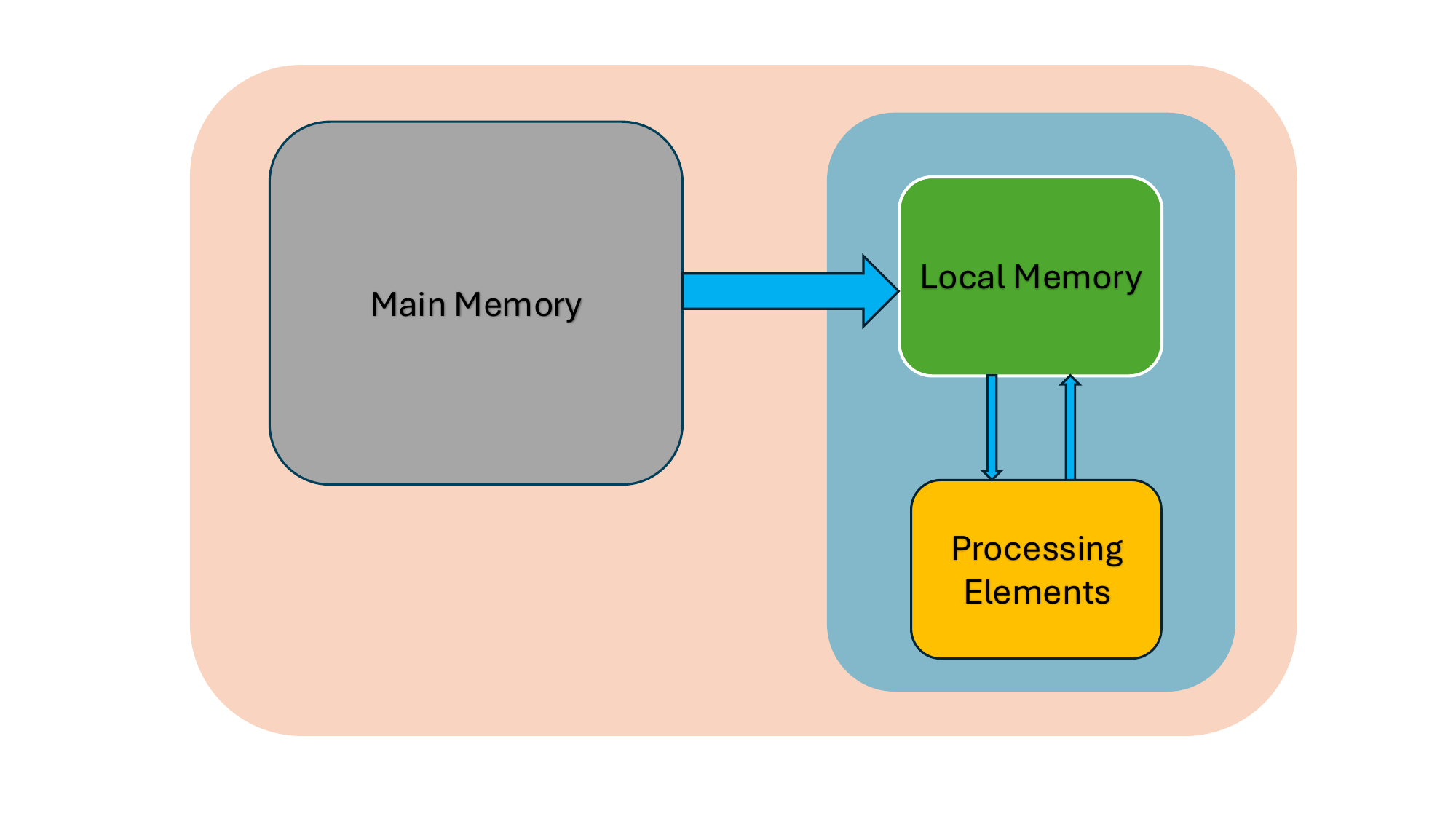}
  \caption{Simple computation model used in Appendix~\ref{sec:energyConsumption}.}
  \label{fig:compute-arch}
\end{figure}

In order to calculate the energy savings netted by the use of our binary stochastic forward-forward (BSFF) algorithm, here we make explicit how memory and multiplication operations scale with network parameters for each of BSFF, CwC-FF \cite{papachristodoulou2024convolutional}, and backprop.
For simplicity, we consider the same architecture in all cases.
This architecture consists of $L$ layers, with each layer containing the following sublayers:
\begin{itemize}
    \item{convolution}
    \item{activation}
    \item{max pooling}
    \item{batch normalization}
\end{itemize}
We assume a simple Von-Neumann computational model, comprising memory units and compute units, as shown abstractly in Figure~\ref{fig:compute-arch}. Although Forward–Forward–based algorithms can benefit from more advanced architectures that exploit multi–level memory hierarchies—since gradient computations are local to each layer—they still conform to the model below.

Specifically, we assume that inputs and weights are fetched from main memory, and that each sublayer has sufficient local memory to store its outputs and weights either for a single sample (one datum per batch and one layer at a time) or for one channel across the entire batch. In general, intermediate tensors can have  dimensions $\bigl(\text{Channels}\,\times\,\text{Height}\,\times\,\text{Width}\bigr)$ or 
$\bigl(\text{Batch Size}\,\times\,\text{Height}\,\times\,\text{Width}\bigr)$ (as $C \sim N$).
This assumption is reasonable, since even a mid‑sized network requires only on the order of 2–3 MB of local memory.

To compute gradients for weight updates, we process each sample independently—reading one sample from memory, computing its gradient, and then averaging over the batch. This approach is broadly applicable, with one caveat for BatchNorm: computing the BatchNorm gradients requires the batch mean and variance, as well as the per‑sample BatchNorm output. However, this does not significantly increase memory traffic, because the batch statistics are computed during the forward pass and can be reused in the backward pass.
\subsection{\textbf{Backpropagation: Number of multiplications and memory accesses}}
\subsubsection{Forward Pass(layer $\ell$)}
\begin{align}
    z^{(\ell)} &= w \ast u^{(\ell-1)} \label{eq:conv} \\
    a^{(\ell)} &= \mathrm{ReLU}(z^{(\ell)}) \label{eq:relu} \\
    \Tilde{u}^{(\ell)} &= \mathrm{max\_pool}\left(a^{(\ell)}\right), \label{eq:Maxpool} \\
    u^{(\ell)} &= \mathrm{BN}\left(\Tilde{u}^{(\ell)}\right). \label{eq:bn}
\end{align}

\begin{algorithm}[H]
\caption{ForwardPassOneLayer}
\label{alg:forward-pass-layer}
\begin{algorithmic}[1]
\item[\algorithmicrequire] Layer index $\ell$, batch size $N$
\item[\algorithmicensure] Batch-normalized outputs $u$ for layer $\ell$

\COMMENT{First Pass: Accumulate Mean \& Variance Stats}
\STATE sum $\gets 0$, \quad sumSq $\gets 0$
\STATE \label{alg:weight-read} \textbf{readFromMainMemory}$(W^{(\ell)})$
\FOR{$n \gets 1$ to $N$}
\IF{$\ell = 1$}
  \STATE \label{alg:input-read-layer1}\textbf{readFromMainMemory}$(\textit{input}_n)$
    \STATE \label{alg:conv-1}$z_n^{(\ell)} \gets \mathrm{Conv}(W^{(\ell)}, \textit{input}_n)$
\ELSE
  \STATE \label{alg:input-read-rest-layer}\textbf{readFromMainMemory}$(u_n^{(\ell-1)})$
    \STATE \label{alg:conv-n}$z_n^{(\ell)} \gets \mathrm{Conv}(W^{(\ell)}, u_n^{(\ell-1)})$
\ENDIF
  \STATE \label{alg:relu-store}$a_n^{(\ell)} \gets \mathrm{ReLU}(z_n^{(\ell)})$, \textbf{writeToMainMemory}$(a_n^{(\ell)})$
  \STATE \label{alg:Maxpool-store}$\Tilde{u}_n^{(\ell)} \gets \mathrm{Maxpool}(a_n^{(\ell)}), \textbf{writeToMainMemory}(\Tilde{u}_n^{(\ell)})$
  \STATE \label{alg:sum}sum $\gets \text{sum} + \text{elementwiseSum}(\Tilde{u}_n^{(\ell)})$
  \STATE \label{alg:sumsq}sumSq $\gets \text{sumSq} + \text{elementwiseSumOfSquares}(\Tilde{u}_n^{(\ell)})$
\ENDFOR
\STATE \label{alg:mean-var-calc}$\mu \gets \frac{\text{sum}}{\textit{totalElements}}$
\STATE \label{alg:mean-var-store}$\sigma^2 \gets \frac{\text{sumSq}}{\textit{totalElements}} - (\textit{mean})^2$
\STATE \label{alg:store-mean-var}\textbf{writeToMainMemory}$(\mu)$, \textbf{writeToMainMemory}$(\sigma^2)$

\COMMENT{Apply BatchNorm using computed stats}
\FOR{$n \gets 1$ to $N$}
  \STATE \label{alg:Maxpool-read}\textbf{readFromMainMemory}$(\Tilde{u}_n^{(\ell)})$
  \STATE \label{alg:BatchNorm-store}$u \gets \mathrm{BatchNorm}(\Tilde{u}_n^{(\ell)}, \mu, \sigma^2)$, \textbf{writeToMainMemory}$(u_n)$
\ENDFOR
\STATE \textbf{return} $u$
\end{algorithmic}
\end{algorithm}

\begin{itemize}
\item \textbf{Total Number of Multiplications} (per layer)

\begin{align*}
\text{MulCount} 
= & \; \underbrace{
  N C^{(\ell-1)} C^{(\ell)} K^2 H^{(\ell-1)} W^{(\ell-1)}
}_{\substack{\text{convolution, Algorithm~\ref{alg:forward-pass-layer},}\\\text{Line~\ref{alg:conv-1}} \text{ or Line~\ref{alg:conv-n}}}} 
\end{align*}
\hfill


  \item \textbf{\textbf{Number of Memory Accesses} (per layer)}

\begin{align*}
\text{MemAccess} 
= & \; \underbrace{
  C^{(\ell-1)}C^{(\ell)}K^2
}_{\substack{\text{weight read, Algorithm~\ref{alg:forward-pass-layer},}\\\text{Line~\ref{alg:weight-read}}}} \\
+ & \; \underbrace{
  N C^{(\ell-1)} H^{(\ell-1)} W^{(\ell-1)}
}_{\substack{\text{input read, Algorithm~\ref{alg:forward-pass-layer},}\\\text{Line~\ref{alg:input-read-layer1} or \ref{alg:input-read-rest-layer}}}} \\
+ & \; \underbrace{
  N C^{(\ell)} H^{(\ell)} W^{(\ell)}
}_{\substack{\text{activation store, Algorithm~\ref{alg:forward-pass-layer},}\\\text{Line~\ref{alg:relu-store}}}} \\
+ & \; \underbrace{
  N C^{(\ell)} H^{(\ell)} W^{(\ell)}
}_{\substack{\text{Maxpool output store, Algorithm~\ref{alg:forward-pass-layer},}\\\text{Line~\ref{alg:Maxpool-store}}}} \\
+ & \; \underbrace{
  N C^{(\ell)} H^{(\ell)} W^{(\ell)}
}_{\substack{\text{Maxpool output read for BatchNorm, } \\ \text{Algorithm~\ref{alg:forward-pass-layer},}\text{Line~\ref{alg:Maxpool-read}}}} \\
+ & \; \underbrace{
  N C^{(\ell)} H^{(\ell)} W^{(\ell)}
}_{\substack{\text{BatchNorm output store, Algorithm~\ref{alg:forward-pass-layer},}\\\text{Line~\ref{alg:BatchNorm-store}}}}
\end{align*}
\end{itemize}

\subsubsection{Backward Pass}
To compute the weight gradient \( \frac{\partial L}{\partial W^{(\ell)}} \) for any layer \( \ell \), the following steps are followed. The gradient of the Maxpool operation involves only an indexing operation through the maximum element in each patch. Therefore, for simplicity, we combine the BatchNorm Jacobian and \( \frac{\partial u^{(\ell)}}{\partial \tilde{u}^{(\ell)}} \) with the Maxpool Jacobian \( \frac{\partial \tilde{u}^{(\ell)}}{\partial a^{(\ell)}} \), and directly express \( \frac{\partial u^{(\ell)}}{\partial a^{(\ell)}} \), as this does not affect our analysis of the number of multiplications and memory accesses.

\begin{enumerate}
\item \textbf{Initialize for the last layer:} 
  \begin{equation*}
  \begin{split}
    \Delta^{(L)}[c,h,w]
    &= \frac{\partial L}{\partial u^{(L)}[c,h,w]}\,
  \end{split}
\end{equation*}
for each channel \(c\) and spatial location \((h,w)\).

\item \textbf{Backward pass for layers $\ell = L$ down to $1$:}
  \begin{enumerate}
    \item \textbf{Compute the gradient w.r.t.\ pre-activation $z^{(\ell)}$:}
\begin{equation*}
  \begin{split}
  \frac{\partial L}{\partial z^{(\ell)}[c,h,w]}
  &=\; \Delta^{(\ell)}[c,h,w] 
       \;\cdot\;
       \Bigl[ 
         \underbrace{\frac{\partial u^{(\ell)}[c,h,w]}
                          {\partial \tilde{u}^{(\ell)}[c,h,w]}}_{\text{BN derivative}}\\
        & \;\cdot\;
         \underbrace{\frac{\partial a^{(\ell)}[c,h,w]}
                          {\partial z^{(\ell)}[c,h,w]}}_{\text{ReLU or BSN derivative}}
       \Bigr]
  \end{split}       
\end{equation*}
Here, $\Delta^{(\ell)}[\,c,\,h,\,w\,]$ is the upstream gradient from the “goodness” function or from higher layers.

    \item \textbf{Compute the weight gradient:}
\begin{equation*}
  \begin{split}
      \frac{\partial L}{\partial W^{(\ell)}[\,c,\;c',\;k_1,\;k_2\,]} 
      \;=\;
      \sum_{h=1}^{H^{(\ell)}} 
      \;\sum_{w=1}^{W^{(\ell)}}
      \;
      \frac{\partial L}{\partial z^{(\ell)}[\,c,\,h,\,w\,]}\\
      \;\cdot\;
      u^{(\ell-1)}\bigl[c',\,\text{fold}(h,k_1),\,\text{fold}(w,k_2)\bigr]
  \end{split}       
\end{equation*}
where $(k_1,k_2)$ index the kernel offsets (size $K\times K$), and \(\text{fold}(\cdot)\) indicates the spatial mapping back to the correct location in $u^{(\ell-1)}$ (depending on stride and padding). 
In practice, you sum over \textit{all} spatial locations $(h,w)$ of $z^{(\ell)}$ that overlap with the kernel position $(k_1,k_2)$ on channel $c'$ of $u^{(\ell-1)}$.

    \item \textbf{Compute the upstream gradient for layer $\ell-1$:}
\begin{equation*}
  \begin{split}
      \Delta^{(\ell-1)}[\,c',\,h',\,w'\,]
      =
      \sum_{c=1}^{C^{(\ell)}}
      \;\sum_{k_1=1}^{K}\;\sum_{k_2=1}^{K}
      \;
      \frac{\partial L}{\partial z^{(\ell)}[\,c,\;h,\;w\,]}\\
      \;\cdot\;
      W^{(\ell)}[\,c,\;c',\;k_1,\;k_2\,].
  \end{split}       
\end{equation*}
Only those elements of $W^{(\ell)}$ that map $(h',w')$ in $u^{(\ell-1)}$ to $(h,w)$ in $z^{(\ell)}$ will be nonzero in this summation (i.e., $C^{(\ell-1)} \times K^2$ effective connections per output channel, if stride=1).
  \end{enumerate}
\end{enumerate}
\begin{itemize}
\item \textbf{Total Number of Multiplications} (per layer)
  \begin{align*}
  = 
    & \underbrace{2NC^{(\ell)}H^{(\ell)}W^{(\ell)}}_{\text{Algorithm \ref{alg:backprop-arbitrary-layer}, Line \ref{alg:backprop-BP:sigma_term_compute}}}
    + \underbrace{NC^{(\ell)}H^{(\ell)}W^{(\ell)}}_{\text{Algorithm \ref{alg:backprop-arbitrary-layer}, Line \ref{alg:backprop-BP:dL_da_compute}}} \\
    & + \underbrace{NC^{(\ell)}H^{(\ell)}W^{(\ell)} C^{(\ell-1)}K^2}_{\text{Algorithm \ref{alg:backprop-arbitrary-layer}, Line \ref{alg:backprop-BP:compute_dL_dW}}} \\
    & + \underbrace{NC^{(\ell)}H^{(\ell)}W^{(\ell)} C^{(\ell-1)}K^2}_{\text{Algorithm \ref{alg:backprop-arbitrary-layer}, Line \ref{alg:backprop-BP:compute_dL_du_prev}}}
  \end{align*}
  \hfill
  \item \textbf{Total number of memory accesses}(Per Layes)
  \begin{align*}
  =
    & \underbrace{2NC^{(\ell)}H^{(\ell)}W^{(\ell)}}_{\text{Algorithm \ref{alg:backprop-arbitrary-layer}, Line \ref{alg:backprop-BP:read_tilde_u} and \ref{alg:backprop-BP:read_dL_du}}} \\
      & + \underbrace{NC^{(\ell)}H^{(\ell)}W^{(\ell)}}_{\text{Algorithm \ref{alg:backprop-arbitrary-layer}, Line \ref{alg:backprop-BP:read_a}}} \\
      & + \underbrace{NC^{(\ell)}C^{(\ell-1)}H^{(\ell-1)}W^{(\ell-1)}}_{\text{Algorithm \ref{alg:backprop-arbitrary-layer}, Line \ref{alg:backprop-BP:read_u_prev}}} \\
    & + \underbrace{2C^{(\ell)}C^{(\ell-1)}K^2}_{\text{Algorithm \ref{alg:backprop-arbitrary-layer}, Line \ref{alg:backprop-BP:read_weight} and \ref{alg:backprop-BP:write_weight}}}
  \end{align*}

  \end{itemize}
\begin{algorithm}[]
  \caption{Backpropagation for an Arbitrary Layer}
  \label{alg:backprop-arbitrary-layer}
  \begin{algorithmic}[1]
    \ENSURE \label{alg:backprop-BP:ensure} 
      Gradients: $\frac{\partial L}{\partial W}, \frac{\partial L}{\partial u^{(\ell-1)}}$.
      
    \FOR{$c^{(\ell)}=1$ \textbf{to} $C^{(\ell)}$} \label{alg:backprop-BP:for_outer}
    
      \STATE \textbf{readFromMainMemory}$\bigl(\Tilde{u}^{(\ell)}[\,c^{(\ell)}]\bigr)$ \label{alg:backprop-BP:read_tilde_u}
      \STATE \textbf{readFromMainMemory}$\bigl(\frac{\partial L}{\partial u^{(\ell)}}\bigr)$ \label{alg:backprop-BP:read_dL_du}
     \STATE $temp \gets \frac{\partial L}{\partial u^{(\ell)}}$
      \STATE $\displaystyle \mu_{grad} \gets \frac{\sum \, temp}{NHW}$ \label{alg:backprop-BP:mu_grad_compute}
      
      \STATE $\displaystyle \bar{\Tilde{u}} \gets \Tilde{u}[\,c^{(\ell)}] \;-\; \mu\bigl[c^{(\ell)}\bigr]$  
      \label{alg:backprop-BP:compute_u_tilde_bar} 
      \COMMENT{// shape [N, H, W]}
    
      \STATE 
      $\displaystyle 
          \Sigma_{term} 
          \gets 
          \frac{\mathrm{mean}\Bigl(\frac{\partial L}{\partial u^{(\ell)}[\,c^{(\ell)}]} *\bar{\Tilde{u}}\Bigr)}
               {\sigma^2\bigl[c^{(\ell)}\bigr]}
      $ \label{alg:backprop-BP:sigma_term_compute}
      \COMMENT{// $NHW$ multiplications}
      
      \STATE 
      $\displaystyle 
          \frac{\partial L}{\partial a^{(\ell)}}[\,c^{(\ell)}] 
          \gets 
          \frac{\gamma[c^{(\ell)}]}{\sqrt{\sigma^2\bigl[c^{(\ell)}\bigr]}} 
          \;\cdot\; 
          \Bigl( 
            temp \;-\; \mu_{grad} 
            \;-\; \bar{\Tilde{u}} * \Sigma_{term} 
          \Bigr)
      $ \label{alg:backprop-BP:dL_da_compute}
      \COMMENT{// $NHW$ multiplications}      
      \STATE \textbf{readFromMainMemory}$\bigl(a^{(\ell)}[\,c^{(\ell)}]\bigr)$ \label{alg:backprop-BP:read_a}
      \STATE 
      $\displaystyle 
          \frac{\partial L}{\partial z}[\,c^{(\ell)}] 
          \gets 
          \frac{\partial L}{\partial a^{(\ell)}}[\,c^{(\ell)}] 
          * 
          \mathbb{I}\Bigl( a^{(\ell)}[\,c^{(\ell)}] > 0 \Bigr)
      $ \label{alg:backprop-BP:compute_dL_dz}
      \COMMENT{// 0 multiplications}
      
      \STATE \COMMENT{// Gradient with respect to Convolution Weights and input $u^{(\ell-1)}$} \label{alg:backprop-BP:grad_conv_input_comment}
      
      \FOR{$c^{(\ell-1)}=1$ \textbf{to} $C^{(\ell-1)}$} \label{alg:backprop-BP:for_inner}
      
        \STATE \textbf{readFromMainMemory}$\bigl(u^{(\ell-1)}[c^{(\ell-1)}]\bigr)$ \label{alg:backprop-BP:read_u_prev}
        \STATE \textbf{readFromMainMemory}$\bigl(W[c^{(\ell)},c^{(\ell-1)}]\bigr)$ \label{alg:backprop-BP:read_weight}
        
        \STATE           
          $\frac{\partial L}{\partial W}[c^{(\ell)},c^{(\ell-1)}] \gets Conv\Bigl(u^{(\ell-1)}[c^{(\ell-1)}], \frac{\partial L}{\partial z}[\,c^{(\ell)}]\Bigr)
          $ \label{alg:backprop-BP:compute_dL_dW}
        \COMMENT{// $NK^2HW$ multiplications}        
\STATE         $W[c^{(\ell)},c^{(\ell-1)}]  \gets W[c^{(\ell)},c^{(\ell-1)}] - \eta \frac{\partial L}{\partial W}[c^{(\ell)},c^{(\ell-1)}] $ \label{alg:backprop-BP:update_weight}
        \STATE \textbf{writeToMainMemory}$\bigl(W[c^{(\ell)},c^{(\ell-1)}]\bigr)$ \label{alg:backprop-BP:write_weight}
        \STATE 
          $\frac{\partial L}{\partial u^{(\ell-1)}}[c^{(\ell-1)}] \gets TransposedConv\Bigl(u^{(\ell-1)}[c^{(\ell-1)}],W[c^{(\ell)},c^{(\ell-1)}]\Bigr)$ \label{alg:backprop-BP:compute_dL_du_prev}
        \COMMENT{// $NK^2HW$ multiplications}
        
      \ENDFOR \label{alg:backprop-BP:end_for_inner}
      
    \ENDFOR \label{alg:backprop-BP:end_for_outer}
    
    \RETURN $\frac{\partial L}{\partial W}$, $\frac{\partial L}{\partial u^{(\ell-1)}}$ \label{alg:backprop-BP:return}
    
  \end{algorithmic}
\end{algorithm}
\subsection{\textbf{Forward Forward Algorithm (Real): Number of Multiplications and Memory Accesses}}
\subsubsection{Forward Pass (layer $\ell$)}
\begin{align}
    z^{(\ell)}[c] 
    &= 
    \sum_{c'} 
    W^{(\ell)}[\,c,\;c'\,] 
    \;\ast\;
    u^{(\ell-1)}[\,c'\,]
    && 
    \text{(convolution)}
    \label{eq:FF_conv}
    \\[6pt]
    a^{(\ell)}[c] 
    &= 
    \mathrm{ReLU}\!\Bigl(z^{(\ell)}[c]\Bigr)
    && 
    \text{(activation)}
    \label{eq:FF_relu}
    \\[6pt]
    \tilde{u}^{(\ell)}[c] 
    &= 
    \mathrm{max\_pool}\!\Bigl(a^{(\ell)}[c]\Bigr)
    && 
    \text{(max pooling)}
    \label{eq:FF_Maxpool}
    \\[6pt]
    u^{(\ell)}[c] 
    &= 
    \mathrm{BN}\!\Bigl(\tilde{u}^{(\ell)}[c]\Bigr)
    && 
    \text{(batch normalization)}
    \label{eq:FF_BatchNorm}
\end{align}

\begin{algorithm}[]
  \caption{ForwardForwardOneLayer (Real Valued Activations)}
  \label{alg:forward-forward-layer}
  \begin{algorithmic}[1]
  
  \item[\algorithmicrequire] 
    Layer index $\ell$, batch size $N$
    \quad
    \label{alg:forward-forward-layer:requires}

  \item[\algorithmicensure] 
    Updated weights $W^{(\ell)}$ and batch-normalized outputs $u^{(\ell)}$
    \quad
    \label{alg:forward-forward-layer:ensures}
  
  \COMMENT{First Pass: Accumulate Mean \& Variance for BN}
  \label{alg:forward-forward-layer:accumulate-mean-variance}
  
  \STATE sum $\gets 0$, \quad sumSq $\gets 0$
  \label{alg:forward-forward-layer:init-sums}
  
  \STATE \textbf{readFromMainMemory}$(W^{(\ell)})$
  \label{alg:forward-forward-layer:read-weights}

  \FOR{$n \gets 1$ to $N$}
  \label{alg:forward-forward-layer:loop-n}
  
    \IF{$\ell = 1$}
    \label{alg:forward-forward-layer:if-first-layer}
    
        \STATE \textbf{readFromMainMemory}$(\textit{input}_n)$
        \label{alg:forward-forward-layer:read-input}
        
        \STATE $z_n^{(\ell)} \gets \mathrm{Conv}\bigl(W^{(\ell)},\,\textit{input}_n\bigr)$
        \label{alg:forward-forward-layer:conv-first-layer}
        \COMMENT{Real conv if first layer, $C[\ell-1]C[\ell]HWK^2$ multiplications}
        
    \ELSE
    \label{alg:forward-forward-layer:else-higher-layer}
    
        \STATE \textbf{readFromMainMemory}$\bigl(u_n^{(\ell-1)}\bigr)$
        \label{alg:forward-forward-layer:read-prev-activ}
        
        \STATE $z_n^{(\ell)} \gets  \mathrm{Conv}\bigl(W^{(\ell)}, u_n^{(\ell-1)}\bigr)$
        \label{alg:forward-forward-layer:conv-binary}
        \COMMENT{Real conv, $K^2HWC[\ell-1]C[\ell]$ multiplications}
        
    \ENDIF
  
    \STATE $a_n^{(\ell)} \gets \mathrm{BSN}\bigl(z_n^{(\ell)}\bigr)$
    \quad \textbf{writeToMainMemory} $\bigl(\Delta_n^{(\ell)}\bigr)$
    \label{alg:forward-forward-layer:bsn-forward}
    \COMMENT{Storing relu gradient instead of $a^{(\ell)}$}

    \STATE $\tilde{u}_n^{(\ell)} \gets \mathrm{Maxpool}\bigl(a_n^{(\ell)}\bigr)$
    \quad \textbf{writeToMainMemory}$\bigl(\tilde{u}_n^{(\ell)}\bigr)$
    \label{alg:forward-forward-layer:maxpool}
    
    \STATE sum   $\gets$ sum + \text{elementwiseSum}$\bigl(\tilde{u}_n^{(\ell)}\bigr)$
    \label{alg:forward-forward-layer:update-sum}
    
    \STATE sumSq $\gets$ sumSq + \text{elementwiseSumOfSquares}$\bigl(\tilde{u}_n^{(\ell)}\bigr)$
    \label{alg:forward-forward-layer:update-sumsq}
    
  \ENDFOR
  
  \STATE $\mu^{(\ell)} \gets \dfrac{\text{sum}}{\textit{totalElements}}$
  \label{alg:forward-forward-layer:compute-mean}

  \STATE $\bigl(\sigma^{(\ell)}\bigr)^2 
         \gets \dfrac{\text{sumSq}}{\textit{totalElements}}
         \;-\;\bigl(\mu^{(\ell)}\bigr)^2$
  \label{alg:forward-forward-layer:compute-variance}
    
\STATE \COMMENT{BN and Weight Updates} \label{alg:forward-forward-layer:local-grad-update}
\FOR{$n \gets 1$ to $N$} \label{alg:forward-forward-layer:ff-second-loop}
  \STATE \textbf{readFromMainMemory}$(\Tilde{u}_n^{(\ell)})$ \label{alg:forward-forward-layer:ff-read-pool}
  \STATE $u_n^{(\ell)} \gets \mathrm{BatchNorm}(\Tilde{u}_n^{(\ell)}, \mu, \sigma^2)$ \label{alg:forward-forward-layer:ff-BatchNorm}
  \STATE \textbf{writeToMainMemory}$(u_n^{(\ell)})$ \label{alg:forward-forward-layer:ff-store-bn}
\ENDFOR

  \IF{$\ell = 1$}
  \label{alg:forward-forward-layer:if-compute-grad-first}
  
    \STATE 
    $\dfrac{\partial L^{(\ell)}}{\partial W^{(\ell)}} 
            \gets \textbf{ComputeGradient}\bigl(\tilde{u}^{(\ell)},{u}^{(\ell)},\Delta^{(\ell)},\,\textit{input}\bigr)$  
    \label{alg:forward-forward-layer:compute-gradient-first}
    
  \ELSE
  \label{alg:forward-forward-layer:else-compute-grad-higher}
  
    \STATE 
    $\dfrac{\partial L^{(\ell)}}{\partial W^{(\ell)}} 
            \gets \textbf{ComputeGradient}\bigl(\tilde{u}^{(\ell)},{u}^{(\ell)},\Delta^{(\ell)},\,u^{(\ell-1)}\bigr)$  
    \label{alg:forward-forward-layer:compute-gradient-higher}
    
  \ENDIF
  
  \STATE 
  $W^{(\ell)} \gets W^{(\ell)} 
    \;-\; \eta \cdot
          \dfrac{\partial L^{(\ell)}}{\partial W^{(\ell)}}$
  \label{alg:forward-forward-layer:update-weights}
    
  \STATE $\textbf{writeToMainMemory}\bigl(W^{(\ell)}\bigr)$
  \label{alg:forward-forward-layer:write-weights}
  
  \end{algorithmic}
\end{algorithm}
\begin{itemize}
    
\item \textbf{Total Number of Multiplications (per layer)}
\begin{align*}
= & \; N \, C^{(\ell-1)}C^{(\ell)}K^2H^{(\ell-1)}W^{(\ell-1)} 
    \quad (\text{Algo \ref{alg:forward-forward-layer}}, \text{Lines~\ref{alg:forward-forward-layer:conv-first-layer} or \ref{alg:forward-forward-layer:conv-binary}}) 
\end{align*}
\hfill
\item \textbf{Number of Memory Accesses (per layer)}
\begin{align*}
= & \; C^{(\ell-1)}C^{(\ell)}K^2 
    \quad (\text{Algo \ref{alg:forward-forward-layer}},\text{Line~\ref{alg:forward-forward-layer:read-weights}}) \\
+ & \; N \, C^{(\ell-1)}H^{(\ell-1)}W^{(\ell-1)} 
    \quad (\text{Algo \ref{alg:forward-forward-layer}},\text{Line~\ref{alg:forward-forward-layer:read-input} or \ref{alg:forward-forward-layer:read-prev-activ}}) \\
+ & \; 2N \, C^{(\ell)}H^{(\ell)}W^{(\ell)} 
    \quad (\text{Algo \ref{alg:forward-forward-layer}},\text{Lines~\ref{alg:forward-forward-layer:bsn-forward} and \ref{alg:forward-forward-layer:maxpool}}) \\
+ & \; N \, C^{(\ell)}H^{(\ell)}W^{(\ell)} 
    \quad (\text{Algo \ref{alg:forward-forward-layer}},\text{Line~\ref{alg:forward-forward-layer:ff-read-pool}}) \\
+ & \; N \, C^{(\ell)}H^{(\ell)}W^{(\ell)} 
    \quad (\text{Algo \ref{alg:forward-forward-layer}},\text{Line~\ref{alg:forward-forward-layer:ff-store-bn}}) \\
+ & \; C^{(\ell)}C^{(\ell-1)}K^2 
    \quad (\text{Algo \ref{alg:forward-forward-layer}},\text{Line~\ref{alg:forward-forward-layer:write-weights}})
\end{align*}
\end{itemize}

\subsubsection{\textbf{Weight Update at layer $\ell$:  $\frac{\partial L^{(\ell)}}{\partial w^{(\ell})}$}}

Here the loss is defined by grouping channels $c$ into subsets $S_j$.
\begin{align}
\frac{\partial L}{\partial 
  z^{(\ell)}}[n,c,h,w]
&=\;
\sum_{j=1}^{\mathrm{N_{class}}}
\;\frac{\partial L}{\partial G_j}
\;\frac{\partial G_j}{\partial u^{(\ell)}[c,n,h,w]} \notag \\ 
&\cdot\frac{\partial u^{(\ell)}[n,c,h,w]}{\partial a^{(\ell)}[n,c,h,w]}
\cdot\;\frac{\partial a^{(\ell)}[n,c,h,w]}{\partial z^{(\ell)}[n,c,h,w]},
\label{eq:DeltaZ_FF}
\\[6pt]
\frac{\partial L}{\partial 
  w^{(\ell)}[\,c,\,c',\,k_1,\,k_2]
}
&=\;
\sum_{n=1}^{N}
\;\sum_{h=1}^{H}
\;\sum_{w=1}^{W}
\;\frac{\partial L}{\partial 
  z^{(\ell)}}[n,c,h,w] * \\
&\Bigl[u^{(\ell-1)}\bigl(n,c',h^{\ast},\,w^{\ast}\bigr)\Bigr] \notag
\end{align}

\begin{algorithm}[]
  \caption{ComputeGradient Subroutine for Real Valued FF Network}
  \label{alg:compute-gradient-FF}
  \begin{algorithmic}[1]
  
  \item[\algorithmicrequire] 
    Batchnorm Input $\Tilde{u}^{(\ell)}$, Batchnorm Output $u^{(\ell)}$, ReLU gradient $\Delta^{(\ell)}$, Previous layer batchnorm output ${u}^{(\ell-1)}$.
    \label{alg:compute-gradient-FF:requires}
    
  \item[\algorithmicensure] 
    $\displaystyle \frac{\partial L^{(\ell)}}{\partial W^{(\ell)}}$
    \label{alg:compute-gradient-FF:ensures}

  \STATE $\displaystyle \frac{\partial L^{(\ell)}}{\partial W^{(\ell)}} \gets 0$
  \label{alg:compute-gradient-FF:init-partial-LW}

  \FOR{$c^{(\ell)} \gets 1$ to $C^{(\ell)}$}
  \label{alg:compute-gradient-FF:loop-c-l}
    \STATE \textbf{readFromMainMemory}$\bigl({u}^{(\ell)}[c^{(\ell)}]\bigr)$
    \label{alg:compute-gradient-FF:read-tilde-u-n}  
    \STATE $\text{group index}:j \gets \frac{c^{(\ell)}}{N_{classes}}$
        \STATE  
        $\frac{\partial G_j}{\partial u^{(\ell)}[c^{(\ell)}]} 
        \gets 
        \frac{2}{|N_j|} 
        \Bigl({u}^{(\ell)}[c^{(\ell)}] \Bigr)$
      \label{alg:compute-gradient-FF:dG-du}
      \COMMENT{// shape $[N,H,W]$}

      \STATE  
        $\displaystyle 
        \frac{\partial L^{(\ell)}}{\partial u^{(\ell)}[c^{(\ell)}]} 
        \gets  
        \frac{\partial G_j}{\partial u[c^{(\ell)}]} 
        \;\ast\; 
        \frac{\partial L^{(\ell)}}{\partial G_{j}}
        $
      \label{alg:compute-gradient-FF:dL-du}
      \COMMENT{// $N \times H \times W$ multiplications}

    \STATE $\displaystyle temp \gets \frac{\partial L^{(\ell)}}{\partial u^{(\ell)}[\,c^{(\ell)}]}$
    \label{alg:compute-gradient-FF:temp-def}
    \COMMENT{// shape [N, H, W]}

    \STATE $\displaystyle \mu_{grad} \gets \frac{\sum temp}{NHW}$
    \label{alg:compute-gradient-FF:mu-grad}

    \STATE $\displaystyle \bar{\Tilde{u}} \gets \Tilde{u}[\,c^{(\ell)}] \;-\; \mu\bigl[c^{(\ell)}\bigr]$
    \label{alg:compute-gradient-FF:a-bar}
    \COMMENT{// shape [N, H, W]}

    \STATE 
      $\displaystyle 
      \Sigma_{term} 
      \gets 
      \frac{\mathrm{mean}\Bigl(\frac{\partial L^{(\ell)}}{\partial u^{(\ell)}[\,c^{(\ell)}]} *\bar{\Tilde{u}}\Bigr)}
           {\sigma^2\bigl[c^{(\ell)}\bigr]}
      $
    \label{alg:compute-gradient-FF:Sigma-term}
    \STATE 
      $\displaystyle 
      \frac{\partial L^{(\ell)}}{\partial a^{(\ell)}}[\,c^{(\ell)}] 
      \gets 
      \frac{\gamma[c^{(\ell)}\bigr]}{\sqrt{\sigma^2\bigl[c^{(\ell)}\bigr]}} 
      \;\cdot\; 
      \Bigl( 
        temp \;-\; \mu_{grad} 
        \;-\; \bar{\Tilde{u}} * \Sigma_{term} 
      \Bigr)
      $
    \label{alg:compute-gradient-FF:dL-da-update}
    \STATE \textbf{readFromMainMemory}$\bigl(\Delta^{(\ell)}[\,c^{(\ell)}]\bigr)$
    \label{alg:compute-gradient-FF:read-Delta-l}

    \STATE 
      $\displaystyle 
      \frac{\partial L^{(\ell)}}{\partial z}[\,c^{(\ell)}] 
      \gets 
      \frac{\partial L^{(\ell)}}{\partial a^{(\ell)}}[\,c^{(\ell)}] 
      \;\cdot\; 
      \Delta^{(\ell)}[\,c^{(\ell)}]
      $
    \label{alg:compute-gradient-FF:dL-dz}
    \COMMENT{// 0 multiplications}

    \FOR{$c^{(\ell-1)} \gets 1$ to $C^{(\ell-1)}$}
    \label{alg:compute-gradient-FF:loop-c-l-1}

        \STATE \textbf{readFromMainMemory}$\bigl(u^{(\ell-1)}[\,c^{(\ell-1)}]\bigr)$
        \label{alg:compute-gradient-FF:read-tilde-u-l-1}

        \STATE           
          $\frac{\partial L^{\ell}}{\partial W}[c^{(\ell)},c^{(\ell-1)}] \gets Conv\Bigl(u^{(\ell-1)}[c^{(\ell-1)}],\frac{\partial L^{\ell}}{\partial z}[\,c^{(\ell)}]\Bigr)
          $ \label{alg:compute-gradient-FF:compute-dL_dW}

    \ENDFOR
    \label{alg:compute-gradient-FF:end-loop-c-l-1}

  \ENDFOR
  \label{alg:compute-gradient-FF:end-loop-c-l}

  \STATE 
    \textbf{return} 
    $\displaystyle \frac{\partial L^{(\ell)}}{\partial W^{(\ell)}}$
  \label{alg:compute-gradient-FF:return-partial-LW}

  \end{algorithmic}
\end{algorithm}

\begin{itemize}
\item \textbf{Total number of Multiplications (per layer)}
\begin{align*}
= &  
     \underbrace{NC^{(\ell)}H^{(\ell)}W^{(\ell)}}_{\text{Algo \ref{alg:compute-gradient-FF}},\text{Line~\ref{alg:compute-gradient-FF:dL-du}}} \\
  & \quad + \; \underbrace{NC^{(\ell)}H^{(\ell)}W^{(\ell)}}_{\text{Algo \ref{alg:compute-gradient-FF}},\text{Line~\ref{alg:compute-gradient-FF:Sigma-term}}}
    + \underbrace{NC^{(\ell)}H^{(\ell)}W^{(\ell)}}_{\text{Algo \ref{alg:compute-gradient-FF}},\text{Line~\ref{alg:compute-gradient-FF:dL-da-update}}}
 \\
& + \; \underbrace{NC^{(\ell)}C^{(\ell-1)}K^2H^{(\ell)}W^{(\ell)}}_{\text{Algo \ref{alg:compute-gradient-FF}},\text{Line~\ref{alg:compute-gradient-FF:compute-dL_dW}}}
\end{align*}
\hfill
\item \textbf{Total Number of Memory Accesses (per layer)}
\begin{align*}
=\, &  \underbrace{NC^{(\ell)}H^{(\ell)}W^{(\ell)}}_{\text{Algo \ref{alg:compute-gradient-FF}},\text{Line~\ref{alg:compute-gradient-FF:read-tilde-u-n}}} \\
&+   \underbrace{NC^{(\ell)}H^{(\ell)}W^{(\ell)}}_{\text{Algo \ref{alg:compute-gradient-FF}},\text{Line~\ref{alg:compute-gradient-FF:read-Delta-l}}}
+  \underbrace{NC^{(\ell)}C^{(\ell-1)}H^{(\ell-1)}W^{(\ell-1)}}_{\text{Algo \ref{alg:compute-gradient-FF}},\text{Line~\ref{alg:compute-gradient-FF:read-tilde-u-l-1}}}
\end{align*}
\end{itemize}

\subsection{\textbf{Forward Forward Algorithm (BSN): Number of Multiplications and Memory Accesses}}

\subsubsection{Forward Pass (layer $\ell$)}


\begin{align}
z^{(\ell)}[c] \;&=\; w^{(\ell)}[c,\cdot] \;\ast\; u^{(\ell-1)}[\cdot]
\label{eq:FF_conv_BSN_new}\\[6pt]
p^{(\ell)}[c,i] \;&=\; \sigma\!\Bigl(z^{(\ell)}[c,i]\Bigr)
\label{eq:FF_sigmoid_BSN_new}\\[6pt]
a^{(\ell)}[c,i] \;&\sim\; \mathrm{Bernoulli}\Bigl(p^{(\ell)}[c,i]\Bigr)
\quad\forall\,c,\,i
\label{eq:FF_activation_BSN_new}\\[6pt]
\tilde{u}^{(\ell)}[c] \;&=\; \mathrm{max\_pool}\bigl(a^{(\ell)}[c]\bigr)
\label{eq:FF_Maxpool_BSN_new}\\[6pt]
u^{(\ell)}[c] \;&=\; \mathrm{BN}\Bigl(\tilde{u}^{(\ell)}[c]\Bigr)
\label{eq:FF_BatchNorm_BSN_new}
\end{align}

Because the BSN activation outputs are in the set $\{0,1\}$
\[
u^{(\ell)}[c] \;\in\; \Bigl\{
   \tfrac{-\,\mu[c]}{\sqrt{\sigma^2[c]}}, \;
   \tfrac{1 - \mu[c]}{\sqrt{\sigma^2[c]}}
\Bigr\},
\]
each channel \(c\) of \(u^{(\ell)}\) can be written using two constants \(\alpha^{(\ell)}[c]\) and \(\delta^{(\ell)}[c]\), namely:

\[
u^{(\ell)}[c] 
\;=\; 
\alpha^{(\ell)}[c]\,\tilde{u}^{(\ell)}[c] 
\;+\;
\delta^{(\ell)}[c],
\]
where
\[
\alpha^{(\ell)}[c] 
\;=\; 
\frac{1}{\sqrt{\bigl(\sigma^{(\ell)}[c]\bigr)^2}}
\quad\text{and}\quad
\delta^{(\ell)}[c]
\;=\; 
\frac{-\,\mu^{(\ell)}[c]}{\sqrt{\bigl(\sigma^{(\ell)}[c]\bigr)^2}}.
\]

Rewriting the pre-activation \(z^{(\ell)}[c]\) to highlight these terms:

\begin{equation}
\begin{split}    
z^{(\ell)}[c] 
&=
\sum_{c^{(\ell-1)}=1}^{C^{(\ell-1)}} 
\Bigl(
  w^{(\ell)}[c,\,c^{(\ell-1)}] \,\alpha^{(\ell-1)}[c^{(\ell-1)}]\\
  &\ast
  u^{(\ell-1)}[c^{(\ell-1)}]
  \;+\;
  w^{(\ell)}[c,\,c^{(\ell-1)}] \,\delta^{(\ell-1)}[c^{(\ell-1)}]
\Bigr).
\label{eqn:preactivation_rewrite_new}
\end{split}
\end{equation}

This equation shows in practice we do not need to calculate BatchNorm; we can just calculate these constants and store them in local memory.
Whenever we need the BatchNorm output, we simply multiply with the Maxpool output first, which is a binary tensor.
Per channel, this amounts to $2K^2$ multiplications, where $K \times K$ is the kernel size.\\

\subsubsection*{\textbf{M-\emph{tiled} units}} \label{ss:M-tiled-units}
If we use \(M\) \emph{tiled} units instead of binary units, then the batch‑normalized outputs take values
\begin{align*}
u^{(\ell)}[c] &\;\in\;\Bigl\{\tfrac{i - \mu[c]}{\sqrt{\sigma^2[c]}}
  \;\Big|\; i = 0,1,\dots,M\Bigr\},
\end{align*}
where \(i\) is an integer representable in \(\log_2 M\) bits.  Therefore,
\begin{align*}
w[c]\,\frac{i - \mu[c]}{\sqrt{\sigma^2[c]}}
  &= i\,\frac{w[c]}{\sqrt{\sigma^2[c]}}
    \;-\;\mu[c]\,\frac{w[c]}{\sqrt{\sigma^2[c]}}.
\end{align*}
Naïvely, this adds \(M K^2\) extra multiplications per input–output channel kernel of size \(K\times K\), but since energy and area scale roughly with the square of operand bit‑width, a multiply between the 32‑bit constant \(\tfrac{w[c]}{\sqrt{\sigma^2[c]}}\) and the \(\log_2 M\)-bit integer \(i\) uses far fewer gates and switching activity than a full \(32\times32\)-bit multiply—making the \(M K^2\) extra operations negligible compared to the usual \(2K^2\) full‑width multiplies per channel for smaller values of $M$.


\begin{algorithm}[]
  \caption{ForwardForwardOneLayer (BSN)}
  \label{alg:forward-forward-layer-BSN}
  \begin{algorithmic}[1]
  
  \item[\algorithmicrequire] 
    Layer index $\ell$, batch size $N$
    \quad
    \label{alg:forward-forward-layer-BSN:requires}

  \item[\algorithmicensure] 
    Updated weights $W^{(\ell)}$ and batch-normalized outputs $u^{(\ell)}$
    \quad
    \label{alg:forward-forward-layer-BSN:ensures}
  
  \COMMENT{First Pass: Accumulate Mean \& Variance for BN}
  \label{alg:forward-forward-layer-BSN:accumulate-mean-variance}
  
  \STATE sum $\gets 0$, \quad sumSq $\gets 0$
  \label{alg:forward-forward-layer-BSN:init-sums}
  
  \STATE \textbf{readFromMainMemory}$(W^{(\ell)})$
  \label{alg:forward-forward-layer-BSN:read-weights}

  \FOR{$n \gets 1$ to $N$}
  \label{alg:forward-forward-layer-BSN:loop-n}
  
    \IF{$\ell = 1$}
    \label{alg:forward-forward-layer-BSN:if-first-layer}
    
        \STATE \textbf{readFromMainMemory}$(\textit{input}_n)$
        \label{alg:forward-forward-layer-BSN:read-input}
        
        \STATE $z_n^{(\ell)} \gets \mathrm{Conv}\bigl(W^{(\ell)},\,\textit{input}_n\bigr)$
        \label{alg:forward-forward-layer-BSN:conv-first-layer}
        \COMMENT{Real conv if first layer, $C[\ell-1]C[\ell]HWK^2$ multiplications}
        
    \ELSE
    \label{alg:forward-forward-layer-BSN:else-higher-layer}
    
        \STATE \textbf{readFromMainMemory}$\bigl(u_n^{(\ell-1)}\bigr)$
        \label{alg:forward-forward-layer-BSN:read-prev-activ}
        
        \STATE $z_n^{(\ell)} \gets \sum_{\,c^{(\ell-1)}\,}
          \Bigl(
              \alpha^{(\ell)}\bigl[c^{(\ell-1)}\bigr] \cdot w^{(\ell)}\bigl[c^{(\ell-1)}\bigr]\,                      
              *\, u_n^{(\ell-1)}\bigl[c^{(\ell-1)}\bigr]
              \;+\;
              w^{(\ell)}\bigl[c^{(\ell-1)}\bigr]\,
              \delta^{(\ell)}\bigl[c^{(\ell-1)}\bigr]
          \Bigr)$
        \label{alg:forward-forward-layer-BSN:conv-binary}
        \COMMENT{Binary conv (since $u^{(\ell-1)}\in\{0,1\}$), $2K^2C[\ell-1]C[\ell]$ multiplications}
        
    \ENDIF
  
    \STATE $a_n^{(\ell)} \gets \mathrm{BSN}\bigl(z_n^{(\ell)}\bigr)$
    \quad \textbf{writeToMainMemory} $\bigl(\Delta_n^{(\ell)}\bigr)$
    \label{alg:forward-forward-layer-BSN:bsn-forward}
    \COMMENT{Storing BSFF gradient instead of $a^{(\ell)}$}

    \STATE $\tilde{u}_n^{(\ell)} \gets \mathrm{Maxpool}\bigl(a_n^{(\ell)}\bigr)$
    \quad \textbf{writeToMainMemory}$\bigl(\tilde{u}_n^{(\ell)}\bigr)$
    \label{alg:forward-forward-layer-BSN:maxpool}
    
    \STATE sum   $\gets$ sum + \text{elementwiseSum}$\bigl(\tilde{u}_n^{(\ell)}\bigr)$
    \label{alg:forward-forward-layer-BSN:update-sum}
    
    \STATE sumSq $\gets$ sumSq + \text{elementwiseSumOfSquares}$\bigl(\tilde{u}_n^{(\ell)}\bigr)$
    \label{alg:forward-forward-layer-BSN:update-sumsq}
    
  \ENDFOR
  
  \STATE $\mu^{(\ell)} \gets \dfrac{\text{sum}}{\textit{totalElements}}$
  \label{alg:forward-forward-layer-BSN:compute-mean}

  \STATE $\bigl(\sigma^{(\ell)}\bigr)^2 
         \gets \dfrac{\text{sumSq}}{\textit{totalElements}}
         \;-\;\bigl(\mu^{(\ell)}\bigr)^2$
  \label{alg:forward-forward-layer-BSN:compute-variance}
    
  \STATE \COMMENT{No need to calculate explicit BN}
  \label{alg:forward-forward-layer-BSN:no-explicit-bn}

  \STATE \COMMENT{Local Gradient Computation \& Weight Update}
  \label{alg:forward-forward-layer-BSN:local-grad-update}

  \IF{$\ell = 1$}
  \label{alg:forward-forward-layer-BSN:if-compute-grad-first}
  
    \STATE 
    $\dfrac{\partial L^{(\ell)}}{\partial W^{(\ell)}} 
            \gets \textbf{ComputeGradient}\bigl(\tilde{u}^{(\ell)},\,\Delta^{(\ell)},\,\textit{input}\bigr)$  
    \label{alg:forward-forward-layer-BSN:compute-gradient-first}
    
  \ELSE
  \label{alg:forward-forward-layer-BSN:else-compute-grad-higher}
  
    \STATE 
    $\dfrac{\partial L^{(\ell)}}{\partial W^{(\ell)}} 
            \gets \textbf{ComputeGradient}\bigl(\tilde{u}^{(\ell)},\,\Delta^{(\ell)},\,u^{(\ell-1)}\bigr)$  
    \label{alg:forward-forward-layer-BSN:compute-gradient-higher}
    
  \ENDIF
  
  \STATE 
  $W^{(\ell)} \gets W^{(\ell)} 
    \;-\; \eta \cdot
          \dfrac{\partial L^{(\ell)}}{\partial W^{(\ell)}}$
  \label{alg:forward-forward-layer-BSN:update-weights}
    
  \STATE $\textbf{writeToMainMemory}\bigl(W^{(\ell)}\bigr)$
  \label{alg:forward-forward-layer-BSN:write-weights}
  
  \end{algorithmic}
\end{algorithm}

  \begin{itemize}
 \item \textbf{Total Number of Multiplications}
\vspace{-0.5em}
\begin{align*}
\text{MulCount} 
=\;& 
\underbrace{
  \begin{cases}
    N\, C^{(\ell-1)}C^{(\ell)}K^2H^{(\ell-1)}W^{(\ell-1)} 
      & \text{if } \ell=1 \\
    N\, \cdot 2\,C^{(\ell-1)}C^{(\ell)}K^2      
      & \text{if } \ell>1
  \end{cases}
}_{\substack{\text{Algo \ref{alg:forward-forward-layer-BSN}},\text{Conv operation at} \\ \text{Line~\ref{alg:forward-forward-layer-BSN:conv-first-layer} (}\ell=1\text{) or}\\ \text{Line~\ref{alg:forward-forward-layer-BSN:conv-binary} (}\ell>1\text{)}}} \\[8pt]
& + \; \underbrace{
  \text{[Gradient Computations at Line~\ref{alg:forward-forward-layer-BSN:compute-gradient-first} or \ref{alg:forward-forward-layer-BSN:compute-gradient-higher}]}
}_{\text{Counted separately}}
\end{align*}
  \item \textbf{Total number of memory accesses}
  \vspace{-0.5em}
  \begin{align*}
  \text{MemAccess}
  =\;& 
  \underbrace{
    \sum_{\ell=2} (N C^{(\ell-1)}H^{(\ell-1)}W^{(\ell-1)})/32
  }_{\substack{\text{Algo \ref{alg:forward-forward-layer-BSN}},\text{read }u^{(\ell-1)}\text{ at line \ref{alg:forward-forward-layer-BSN:read-prev-activ}}\\(\ell>1\text{ case})}}\\
  &+\underbrace{
     (N C^{(\ell-1)}H^{(\ell-1)}W^{(\ell-1)})
  }_{\substack{\text{Algo \ref{alg:forward-forward-layer-BSN}},\text{read }u^{(\ell-1)}\text{ at line \ref{alg:forward-forward-layer-BSN:read-input}}\\(\ell=1\text{ case})}}\\&+ \underbrace{
    \sum_{\ell}C^{(\ell-1)}C^{(\ell)}K^2
  }_{\substack{\text{Algo \ref{alg:forward-forward-layer-BSN}}, \text{read }W^{(\ell)}\text{ at line \ref{alg:forward-forward-layer-BSN:read-weights}}\\(\text{weights per layer})}}
  \\
  &+ \underbrace{
   \sum_{\ell} (N C^{(\ell)}H^{(\ell)}W^{(\ell)})
  }_{\substack{\text{Algo \ref{alg:forward-forward-layer-BSN}}, \text{store }\Delta^{(\ell)}\text{ at line \ref{alg:forward-forward-layer-BSN:bsn-forward}}}}\\
  &+ \underbrace{
   \sum_{\ell} ( N C^{(\ell)}H^{(\ell)}W^{(\ell)})/32
  }_{\substack{\text{Algo \ref{alg:forward-forward-layer-BSN}}, \text{store Maxpool output}\text{ at line \ref{alg:forward-forward-layer-BSN:maxpool}})}}\\
  &+ \underbrace{
    \sum_{\ell} C^{(\ell-1)}C^{(\ell)}K^2
  }_{\substack{\text{Algo \ref{alg:forward-forward-layer-BSN}}, \text{write updated }W^{(\ell)}\text{ at line \ref{alg:forward-forward-layer-BSN:write-weights}}}}
  \end{align*}
 \end{itemize}

\begin{algorithm}[]
  \caption{ComputeGradient Subroutine for BSN Network}
  \label{alg:compute-gradient-BSN}
  \begin{algorithmic}[1]
  
  \item[\algorithmicrequire] 
    Batchnorm Input $\Tilde{u}^{(\ell)}$, BSFF gradient $\Delta^{(\ell)}$, Previous layer batchnorm input $\Tilde{u}^{(\ell-1)}$.
    \label{alg:compute-gradient-BSN:requires}
    
  \item[\algorithmicensure] 
    $\displaystyle \frac{\partial L^{(\ell)}}{\partial W^{(\ell)}}$
    \label{alg:compute-gradient-BSN:ensures}

  \STATE $\displaystyle \frac{\partial L^{(\ell)}}{\partial W^{(\ell)}} \gets 0$
  \label{alg:compute-gradient-BSN:init-partial-LW}

  \FOR{$c^{(\ell)} \gets 1$ to $C^{(\ell)}$}
  \label{alg:compute-gradient-BSN:loop-c-l}
  
    \STATE \textbf{readFromMainMemory}$\bigl(\Tilde{u}^{(\ell)}\bigr)$
    \label{alg:compute-gradient-BSN:read-tilde-u-n}

    \STATE $\text{group index}:j \gets \frac{c^{(\ell)}}{N_{classes}}$

      \STATE  
        $\displaystyle 
        \frac{\partial G_j}{\partial u^{(\ell)}[c^{(\ell)}]} 
        \gets 
        \frac{2}{|N_j|} 
        \Bigl(\alpha^{(\ell)}\,\Tilde{u}^{(\ell)}[c^{(\ell)}] 
              + 
              \delta^{(\ell)}\Bigr)$
      \label{alg:compute-gradient-BSN:dG-da}

      \STATE  
        $\displaystyle 
        \frac{\partial L^{(\ell)}}{\partial u^{(\ell)}[c^{(\ell)}]} 
        \gets  
        \frac{\partial G_j}{\partial u[c^{(\ell)}]} 
        \;\ast\; 
        \frac{\partial L^{(\ell)}}{\partial G_{j}}
        $
      \label{alg:compute-gradient-BSN:dL-da}
      \COMMENT{// $\sim 0$ multiplications as $\tilde{u} \in \{0,1\}$ and $\alpha$ can be taken outside per channel}

    \STATE $\displaystyle temp \gets \frac{\partial L^{(\ell)}}{\partial u^{(\ell)}[\,c^{(\ell)}]}$
    \label{alg:compute-gradient-BSN:temp-def}
    \COMMENT{// shape [N, H, W]}

    \STATE $\displaystyle \mu_{grad} \gets \frac{\sum temp}{NHW}$
    \label{alg:compute-gradient-BSN:mu-grad}

    \STATE $\displaystyle \bar{\Tilde{u}} \gets \Tilde{u}[\,c^{(\ell)}] \;-\; \mu\bigl[c^{(\ell)}\bigr]$
    \label{alg:compute-gradient-BSN:a-bar}
    \COMMENT{// shape [N, H, W]}

    \STATE 
      $\displaystyle 
      \Sigma_{term} 
      \gets 
      \frac{\mathrm{mean}\Bigl(\frac{\partial L^{(\ell)}}{\partial u^{(\ell)}[\,c^{(\ell)}]} * \bar{\Tilde{u}}\Bigr)}
           {\sigma^2\bigl[c^{(\ell)}\bigr]}
      $
    \label{alg:compute-gradient-BSN:Sigma-term}
    \COMMENT{// $\sim 0$  multiplications}

    \STATE 
      $\displaystyle 
      \frac{\partial L^{(\ell)}}{\partial a^{(\ell)}}[\,c^{(\ell)}] 
      \gets 
      \frac{\gamma[c^{(\ell)}\bigr]}{\sqrt{\sigma^2\bigl[c^{(\ell)}\bigr]}} 
      \;\cdot\; 
      \Bigl( 
        temp \;-\; \mu_{grad} 
        \;-\; \bar{\Tilde{u}} * \Sigma_{term} 
      \Bigr)
      $
    \label{alg:compute-gradient-BSN:dL-da-update}
    \COMMENT{// $\sim 0$  multiplications}

    \STATE \textbf{readFromMainMemory}$\bigl(\Delta^{(\ell)}[\,c^{(\ell)}]\bigr)$
    \label{alg:compute-gradient-BSN:read-Delta-l}

    \STATE 
      $\displaystyle 
      \frac{\partial L^{(\ell)}}{\partial z}[\,c^{(\ell)}] 
      \gets 
      \frac{\partial L^{(\ell)}}{\partial a^{(\ell)}}[\,c^{(\ell)}] 
      \;\cdot\; 
      \Delta^{(\ell)}[\,c^{(\ell)}]
      $
    \label{alg:compute-gradient-BSN:dL-dz}
    \COMMENT{// NHW multiplications}

    \FOR{$c^{(\ell-1)} \gets 1$ to $C^{(\ell-1)}$}
    \label{alg:compute-gradient-BSN:loop-c-l-1}

      \IF{$\ell = 1$}
      \label{alg:compute-gradient-BSN:if-first-layer}

        \STATE \textbf{readFromMainMemory}$\bigl(\textit{input}[\,c^{(\ell-1)}]\bigr)$
        \label{alg:compute-gradient-BSN:read-input-c-l-1}
        \STATE           
          $\frac{\partial L^{\ell}}{\partial W}[c^{(\ell)},c^{(\ell-1)}]\gets Conv\Bigl({input}[c^{(\ell-1)}],\frac{\partial L^{\ell}}{\partial z}[\,c^{(\ell)}]\Bigr)
          $ \label{alg:compute-gradient-BSN:compute-dL_dW_1}
      \ELSE
      \label{alg:compute-gradient-BSN:else-higher-layer}

        \STATE \textbf{readFromMainMemory}$\bigl(\Tilde{u}^{(\ell-1)}[\,c^{(\ell-1)}]\bigr)$
        \label{alg:compute-gradient-BSN:read-tilde-u-l-1}
        \STATE           
          $\frac{\partial L^{\ell}}{\partial W}[c^{(\ell)},c^{(\ell-1)}] \gets Conv\Bigl(\alpha\bigl[c^{(\ell-1)}\bigr] \tilde{u}^{\ell-1}[c^{\ell-1}+\delta\bigl[c^{(\ell-1)}\bigr] ],\frac{\partial L^{\ell}}{\partial z}[\,c^{(\ell)}]\Bigr)
          $ \label{alg:compute-gradient-BSN:compute-dL_dW}        

      \ENDIF

    \ENDFOR
    \label{alg:compute-gradient-BSN:end-loop-c-l-1}

  \ENDFOR
  \label{alg:compute-gradient-BSN:end-loop-c-l}

  \STATE 
    \textbf{return} 
    $\displaystyle \frac{\partial L^{(\ell)}}{\partial W^{(\ell)}}$
  \label{alg:compute-gradient-BSN:return-partial-LW}

  \end{algorithmic}
\end{algorithm}


\subsubsection{\textbf{Weight Update at layer $\ell$:  $\frac{\partial L^{(\ell)}}{\partial w^{(\ell})}$}}

 \begin{itemize}

\item \textbf{Total Number of Multiplications}
\vspace{-0.5em}
\begin{align*}
\text{MulCount} 
=& \underbrace{
 NC^{(\ell)}H^{(\ell)}W^{(\ell)} 
}_{\substack{\text{Algo \ref{alg:compute-gradient-BSN}} ,\text{Line~\ref{alg:compute-gradient-BSN:dL-dz}}}} \\
+&\
\underbrace{
  NC^{(\ell)}H^{(\ell)}W^{(\ell)}\,C^{(\ell-1)}K^2
}_{\substack{\text{Algo \ref{alg:compute-gradient-BSN}} ,\text{Real multiplications at Line~\ref{alg:compute-gradient-BSN:compute-dL_dW_1}}}} \\
+&\;
\underbrace{
  \sum_{\ell=2} 2NC^{(\ell)}C^{(\ell-1)}K^2
}_{\substack{\text{Algo \ref{alg:compute-gradient-BSN}} ,\text{Conv binary multiplications at Line~\ref{alg:compute-gradient-BSN:compute-dL_dW} for binary case}}} \\
\end{align*}
\hfill
\item \textbf{Total number of memory accesses}
\vspace{-0.5em}
\begin{align*}
\text{MemAccess}
=\;&
\underbrace{
  \sum_{\ell} NC^{(\ell)}H^{(\ell)}W^{(\ell)}/32
}_{\substack{\text{Algo \ref{alg:compute-gradient-BSN}} ,\text{Line~\ref{alg:compute-gradient-BSN:read-tilde-u-n}}} }\\
+&
\underbrace{
  \sum_{\ell} NC^{(\ell)}H^{(\ell)}W^{(\ell)}
}_{\substack{\text{Algo \ref{alg:compute-gradient-BSN}} ,\text{Line~\ref{alg:compute-gradient-BSN:read-Delta-l}}} } \\
+&\;\underbrace{
  \sum_{\ell=2} NC^{(\ell-1)}C^{(\ell)}H^{(\ell-1)}W^{(\ell-1)}/32
}_{\substack{\text{Algo \ref{alg:compute-gradient-BSN}} ,\text{Line~\ref{alg:compute-gradient-BSN:read-tilde-u-l-1}}} }\\
+&\underbrace{
 NC^{(\ell-1)}C^{(\ell)}H^{(\ell-1)}W^{(\ell-1)}
}_{\substack{\text{Algo \ref{alg:compute-gradient-BSN}} ,\text{Line~\ref{alg:compute-gradient-BSN:read-input-c-l-1}}} }
\end{align*}
 \end{itemize}

\subsubsection*{\textbf{No–BatchNorm Case}}\label{ss:no-batchnorm}
Above pseudocode assumes BatchNorm. Without BatchNorm, one can compute the gradient for each sample independently and then average, removing the need to reread inputs for every output channel. Under the local‑memory constraint (only a tensor of shape \(NHW\) or \(CHW\) may reside in fast memory), this per‑sample independence reduces the dominant storage term
$NC^2HW \;\longrightarrow\; NCHW$. Moreover, in the \textbf{BGBSFF} variant, all gradient tensors are binary,hence eliminating the \(2C K^2\) real‑valued multiplications per channel. In fact since gradient tensors $  \frac{\partial G_j}{\partial u[c^{(\ell)}]} $ and $\frac{\partial L^{(\ell)}}{\partial G_{j}}$ are binary, even for first layer despite being input full precision all multiplications are turned into indexing and addition giving further orders of saving.

\section{Supplementary:}\label{sec:supplementary}

\subsection{\textbf{Code Availability}}

The code used in our experiments can be accessed at \href{https://anonymous.4open.science/r/CWCC_BSFF-3DF8/README.md}{this repository}.

\subsection{\textbf{Architecture Details}}
\noindent The same architectures are used as in the \cite{papachristodoulou2024convolutional} paper.

\subsection*{For MNIST, FMNIST, and CIFAR‑10}
\begin{itemize}
  \item \textbf{Layer1}: \texttt{Conv2D (Out Channels = 20)} $\to$ \texttt{BSN} $\to$ \texttt{BatchNorm2D}
  \item \textbf{Layer2}: \texttt{GroupConv2D (Out Channels = 80)} $\to$ \texttt{BSN} $\to$ \texttt{MaxPool} $\to$ \texttt{BatchNorm2D}
  \item \textbf{Layer3}: \texttt{Conv2D (Out Channels = 240)} $\to$ \texttt{BSN} $\to$ \texttt{BatchNorm2D}
  \item \textbf{Layer4}: \texttt{GroupConv2D (Out Channels = 480)} $\to$ \texttt{BSN} $\to$ \texttt{MaxPool} 
  \item \textbf{Layer5}: \texttt{Softmax Classifier}
\end{itemize}

\subsection*{For CIFAR‑100}
\begin{itemize}
  \item \textbf{Layer1}: \texttt{Conv2D (Out Channels = 60)} $\to$ \texttt{BSN} $\to$ \texttt{BatchNorm2D}
  \item \textbf{Layer2}: \texttt{GroupConv2D (Out Channels = 120)} $\to$ \texttt{BSN} $\to$ \texttt{MaxPool} $\to$ \texttt{BatchNorm2D}
  \item \textbf{Layer3}: \texttt{Conv2D (Out Channels = 240)} $\to$ \texttt{BSN} $\to$ \texttt{BatchNorm2D}
  \item \textbf{Layer4}: \texttt{GroupConv2D (Out Channels = 400)} $\to$ \texttt{BSN} $\to$ \texttt{MaxPool} $\to$ \texttt{BatchNorm2D}
  \item \textbf{Layer5}: \texttt{Conv2D (Out Channels = 800)} $\to$ \texttt{BSN} $\to$ \texttt{BatchNorm2D}
  \item \textbf{Layer6}: \texttt{GroupConv2D (Out Channels = 1600)} $\to$ \texttt{BSN} $\to$ \texttt{MaxPool} 
  \item \textbf{Layer7}: \texttt{Softmax Classifier}
\end{itemize}

\subsection{\textbf{Hyperparameter Details}}
Table~\ref{tab:bsff_hyperparams} lists the hyperparameters used in our experiments.\begin{table*}[!t]
\centering
\footnotesize
\begin{tabular}{l c c c c l l}
\toprule
Dataset  & BSFF & LR$_1$ (Conv) & LR$_2$ (Softmax) & Batch & Optimizer & Start–End Epochs \\ 
         & (Units) &               &                  & Size  &           & (layer‑wise)     \\ 
\midrule
MNIST   & 1 & 5e-4 & 5e-3 & 128 & Adam & \texttt{[[0,5],[0,10],[0,15],[0,20],[0,100]]} \\
MNIST   & 2 & 1e-3 & 1e-3 & 128 & Adam & \texttt{[[0,5],[0,10],[0,15],[0,20],[0,100]]} \\
MNIST   & 3 & 1e-3 & 1e-3 & 128 & Adam & \texttt{[[0,5],[0,10],[0,15],[0,20],[0,100]]} \\
MNIST   & 7 & 1e-3 & 1e-3 & 128 & Adam & \texttt{[[0,5],[0,10],[0,15],[0,20],[0,100]]} \\
FMNIST  & 1 & 1e-4 & 1e-3 & 128 & Adam & \texttt{[[0,20],[0,30],[0,40],[0,60],[0,120]]} \\
FMNIST  & 2 & 1e-3 & 1e-3 & 128 & Adam & \texttt{[[0,20],[0,30],[0,40],[0,60],[0,120]]} \\
FMNIST  & 3 & 1e-3 & 1e-3 & 128 & Adam & \texttt{[[0,20],[0,30],[0,40],[0,60],[0,120]]} \\
FMNIST  & 7 & 1e-3 & 1e-3 & 128 & Adam & \texttt{[[0,20],[0,30],[0,40],[0,60],[0,120]]} \\
CIFAR10 & 1 & 1e-3 & 1e-3 & 128 & Adam & \texttt{[[0,25],[0,35],[0,50],[0,75],[0,150]]} \\
CIFAR10 & 2 & 1e-3 & 1e-3 & 128 & Adam & \texttt{[[0,25],[0,35],[0,50],[0,75],[0,150]]} \\
CIFAR10 & 3 & 1e-3 & 1e-3 & 128 & Adam & \texttt{[[0,25],[0,35],[0,50],[0,75],[0,150]]} \\
CIFAR10 & 7 & 1e-3 & 1e-3 & 128 & Adam & \texttt{[[0,25],[0,35],[0,50],[0,75],[0,150]]} \\
\bottomrule
\end{tabular}
\caption{Hyperparameter settings for BSFF experiments on MNIST, FMNIST, and CIFAR‑10.}
\label{tab:bsff_hyperparams}
\end{table*}

\subsection{\textbf{Backpropagation Results}}
Table~\ref{tab:results_bp} presents the results for the FMNIST and CIFAR-10 datasets.
We report the accuracy achieved by backpropagation using the same four-layer networks as in our CwC-FF and BGBSFF models, to enable an apples-to-apples comparison of energy consumption (see Appendix C).
However, we emphasize that backpropagation can generally achieve over $90\%$ accuracy on CIFAR-10 with more advanced architectures.
While it would be ideal to compute the energy costs for such state-of-the-art backpropagation-based results, doing so would make the comparison with our own results increasingly complicated and hypothetical.

\begin{table*}[!t]
\centering
\footnotesize
\begin{tabular}{lcccccc}
\toprule
Dataset  & Backprop      & CwCFF         & CwCFF BGBSFF:1  & BGBSFF:2        & BGBSFF:3        & BGBSFF:7        \\
\midrule
FMNIST   & $92.98 \pm 0.27$ & $91.36 \pm 0.22$ & $77.47 \pm 0.90$ & $87.6  \pm 0.23$ & $88.4  \pm 0.37$ & $89.5  \pm 0.27$ \\
CIFAR10  & $81.76 \pm 0.67$ & $76.32 \pm 0.30$ & $53.5  \pm 0.76$ & $63.6  \pm 0.39$ & $68.8  \pm 0.32$ & $72.4  \pm 0.18$ \\
\bottomrule
\end{tabular}
\caption{Test accuracies (\%) on FMNIST and CIFAR‑10 for various methods.}
\label{tab:results_bp}
\end{table*}

\subsection{\textbf{CIFAR-100  Results (Test Accuracies)}}

Table~\ref{tab:cifar100_prelim} reports the mean and standard deviation of test accuracies over 3 runs on CIFAR-100.  
These results were obtained using goodness prediction for training convolutional layers.  
We observe a 1–2\% accuracy improvement when the final-layer features are passed directly to the Softmax classifier.  
Further tuning of hyperparameters is expected to improve BSFF performance.While the absolute test accuracy of BSFF on CIFAR-100 (43--44\%) is lower than that of standard backpropagation-based models (e.g., $\sim$57\% with backprop in our baseline), it is important to contextualize this result. BSFF operates under significant constraints: (i) it eliminates backward gradient computation entirely, (ii) it uses low-bit binary stochastic neurons throughout training and inference, and (iii) it trains layers locally via a forward-only objective. Despite these constraints, BSFF is able to generalize across 100 fine-grained classes, which is non-trivial. These results underscore the promise of BSFF as a scalable, hardware-efficient alternative to traditional backpropagation, particularly in resource-constrained environments.

\begin{table*}[!t]
\centering
\footnotesize
\begin{tabular}{lcccc}
\toprule
\textbf{Dataset} & \textbf{Backprop} & \textbf{CwCFF FP} & \textbf{BSFF:3} & \textbf{BSFF:7} \\
\midrule
CIFAR100 & $57.25 \pm 0.43$ & $49.83 \pm 0.37$ & $38.7 \pm 0.26$ & $43.6 \pm 0.41$ \\
\bottomrule
\end{tabular}
\caption{Preliminary test accuracies on CIFAR-100 using different training methods.}
\label{tab:cifar100_prelim}
\end{table*}

\subsection{\textbf{Error Convergence Plots}}
\begin{figure*}[t]
    \centering
    \includegraphics[width=0.32\textwidth]{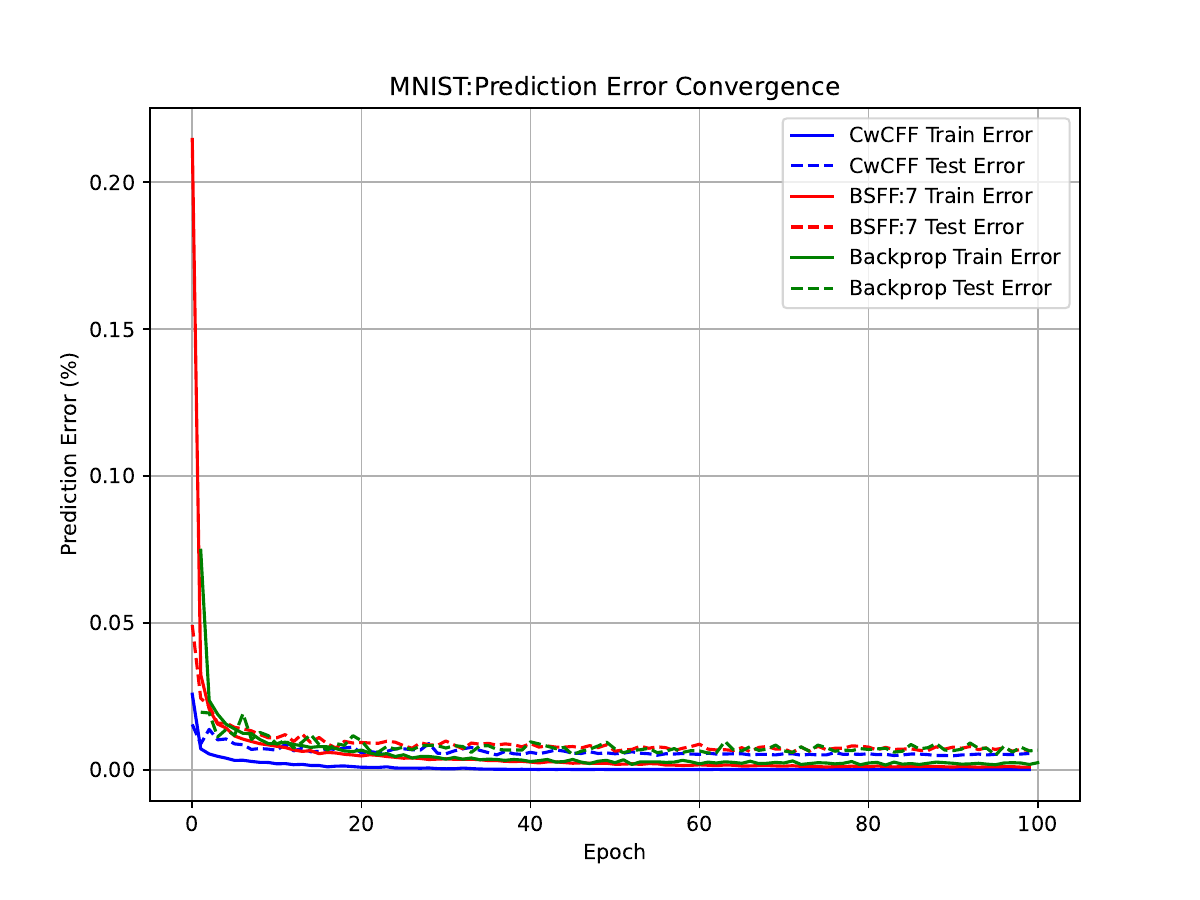} \hfill
    \includegraphics[width=0.32\textwidth]{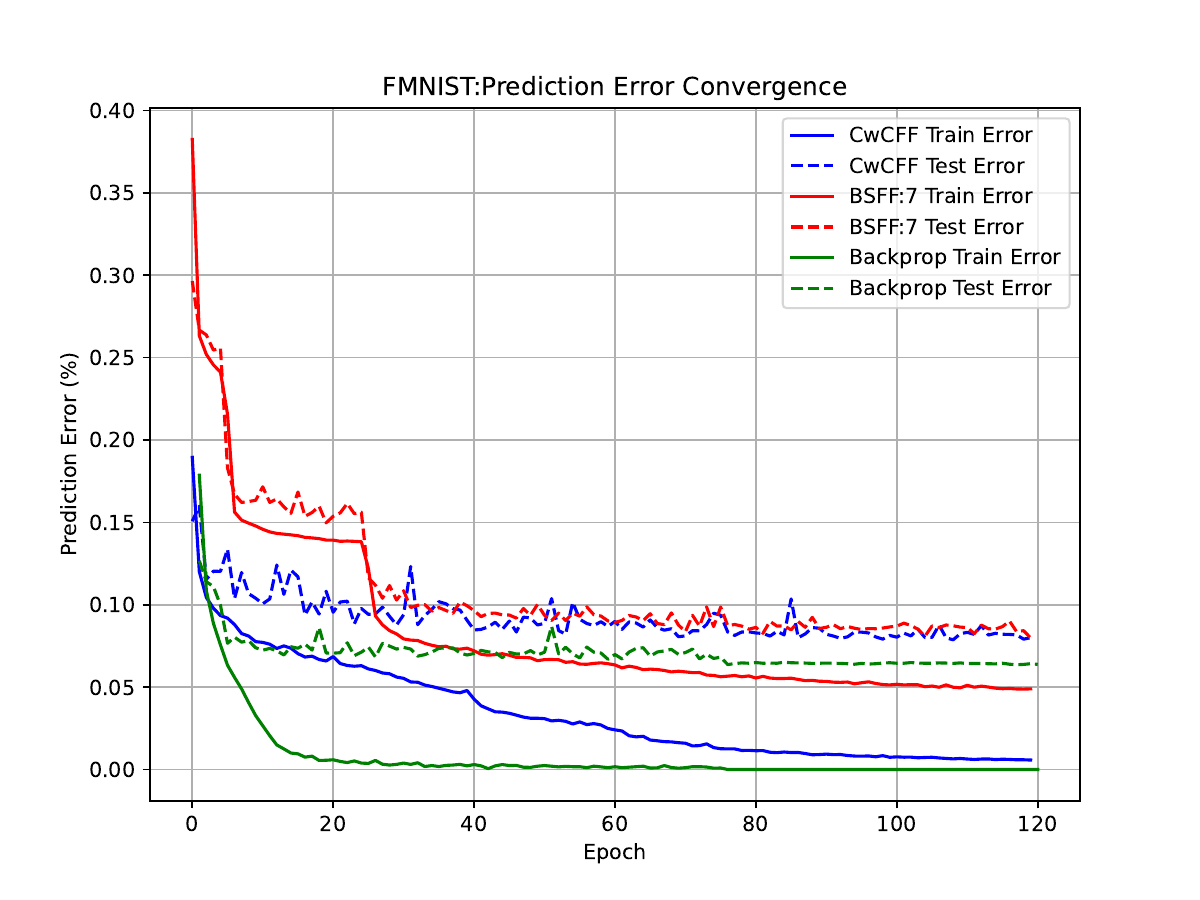} \hfill
    \includegraphics[width=0.32\textwidth]{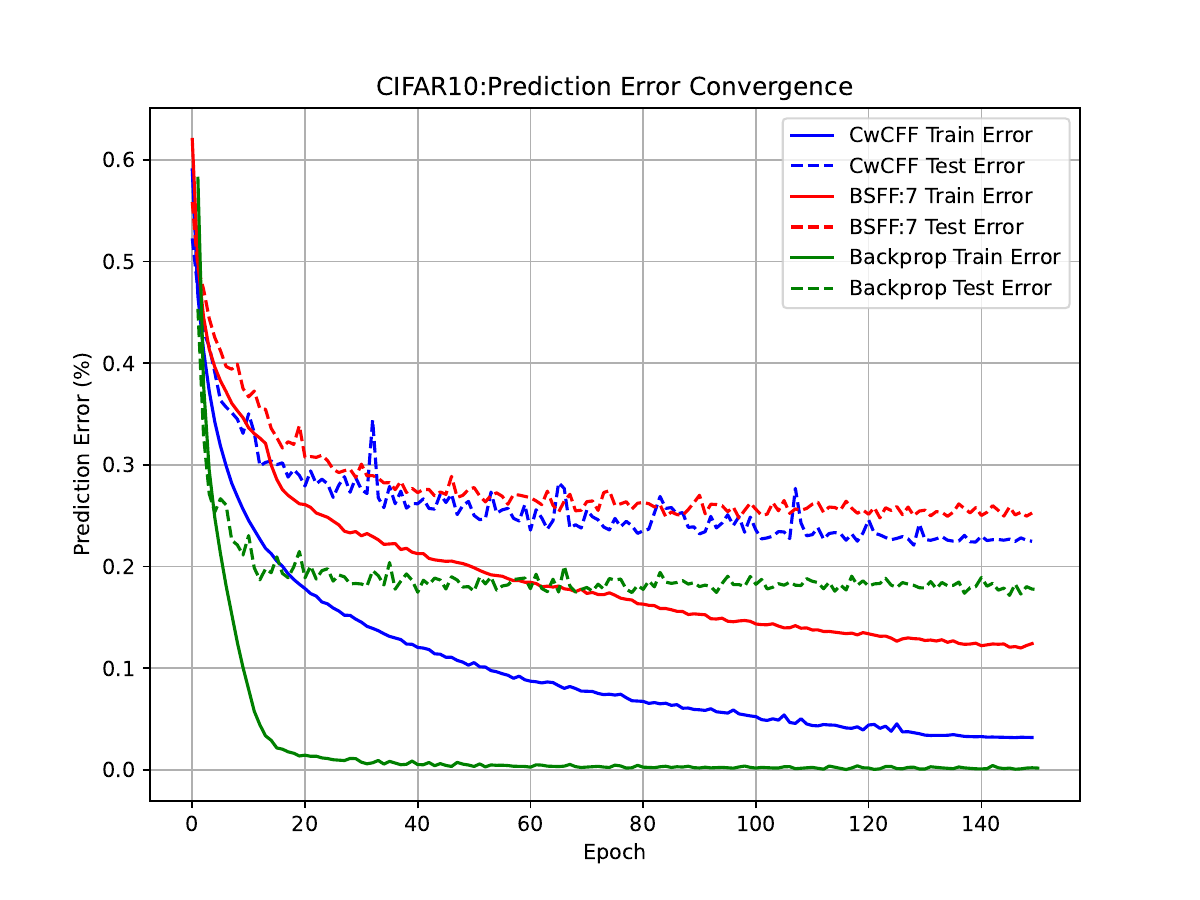}
    \caption{Prediction error convergence plots for MNIST, FMNIST, and CIFAR-10 (from left to right).}
    \label{fig:error-convergence}
\end{figure*}

The prediction error convergence results for different datasets are shown in Figure~\ref{fig:error-convergence}.


\end{document}